\documentclass[journal]{IEEEtran}

\usepackage{amsfonts}
\usepackage{balance}
\usepackage{booktabs}
\usepackage{amsmath}
\usepackage{tabularx}
\usepackage{subfigure}
\usepackage{graphicx}
\usepackage{multirow}
\usepackage{rotating}
\usepackage{algorithmic}
\usepackage{url}

\usepackage{amsthm}
\newtheorem{mydef}{Definition}

\begin{document}

\title{Locally Weighted Ensemble Clustering}

\hyphenation{huangdonghere}
\hyphenation{changdongwang}

\author{Dong~Huang,~\IEEEmembership{Member,~IEEE,}
        Chang-Dong~Wang,~\IEEEmembership{Member,~IEEE,}
        and~Jian-Huang~Lai,~\IEEEmembership{Senior Member,~IEEE}
\thanks{The authors would like to thank the anonymous reviewers for their valuable comments and suggestions. This project was supported by National Key Research and Development Program of China (2016YFB1001003), NSFC (61602189, 61502543 \& 61573387), the PhD Start-up Fund of Natural Science Foundation of Guangdong Province, China (2016A030310457 \& 2014A030310180), and Guangdong Natural Science Funds for Distinguished Young Scholar
(2016A030306014).}
\thanks{Dong Huang is with the College of Mathematics and Informatics, South China Agricultural University, Guangzhou, China. E-mail: huangdonghere@gmail.com.}
\thanks{Chang-Dong Wang and Jian-Huang Lai are
with the School of Data and Computer Science,
Sun Yat-sen University, Guangzhou, China, and also with Guangdong Key Laboratory of Information Security Technology, Guangzhou, China, and also with Key Laboratory of Machine Intelligence and Advanced Computing, Ministry of Education, China.
E-mail: changdongwang@hotmail.com, stsljh@mail.sysu.edu.cn.}
}

\markboth{IEEE Transactions on Cybernetics}%
{Shell \MakeLowercase{\textit{Huang et al.}}: Locally Weighted Ensemble Clustering}

\maketitle

\begin{abstract}
Due to its ability to combine multiple base clusterings into a probably better and more robust clustering, the ensemble clustering technique has been attracting increasing attention in recent years. Despite the significant success, one limitation to most of the existing ensemble clustering methods is that they generally treat all base clusterings equally regardless of their reliability, which makes them vulnerable to low-quality base clusterings. Although some efforts have been made to (globally) evaluate and weight the base clusterings, yet these methods tend to view each base clustering as an individual and neglect the local diversity of clusters inside the same base clustering. It remains an open problem how to evaluate the reliability of clusters and exploit the local diversity in the ensemble to enhance the consensus performance, especially in the case when there is no access to data features or specific assumptions on data distribution. To address this, in this paper, we propose a novel ensemble clustering approach based on ensemble-driven cluster uncertainty estimation and local weighting strategy. In particular, the uncertainty of each cluster is estimated by considering the cluster labels in the entire ensemble via an entropic criterion. A novel ensemble-driven cluster validity measure is introduced, and a locally weighted co-association matrix is presented to serve as a summary for the ensemble of diverse clusters. With the local diversity in ensembles exploited, two novel consensus functions are further proposed. Extensive experiments on a variety of real-world datasets demonstrate the superiority of the proposed approach over the state-of-the-art.
\end{abstract}

\begin{IEEEkeywords}
Ensemble clustering, Consensus clustering, Cluster uncertainty estimation, Local weighting.
\end{IEEEkeywords}

\IEEEpeerreviewmaketitle

\section{Introduction}

\IEEEPARstart{D}{ata} clustering is a fundamental yet still very challenging problem in the field of data mining and machine learning \cite{jain10_survey}. The purpose of it is to discover the inherent structures of a given dataset and partition the dataset into a certain number of homogeneous groups, i.e., clusters. During the past few decades, a large number of clustering algorithms have been developed by exploiting various techniques \cite{xu93_rpcl,ng2002spectral,frey07_ap,wu09_kdd,WangL11_pr,TSMCC12,svstream13,meap13,wang14_tcyb,li2015dias,liu15_icdm,yang15,bao15,wang16_tkde,kumar16_tcyb}. Each clustering algorithm has its advantages as well as its drawbacks, and may perform well for some specific applications. There is no single clustering algorithm that is capable of dealing with all types of data structures and cluster shapes. Given a data set, different clustering algorithms, or even the same algorithm with different initializations or parameters, may lead to different clustering results. However, without prior knowledge, it is extremely difficult to decide which algorithm would be the appropriate one for a given clustering task. Even with the clustering algorithm given, it may still be difficult to find proper parameters for it.

Different clusterings produced by different algorithms (or the same algorithm with different initializations and parameters) may reflect different perspectives of the data. To exploit the complementary and rich information in multiple clusterings, the ensemble clustering technique has emerged as a powerful tool for data clustering and has been attracting increasing attention in recent years \cite{strehl02,cristofor02,fern04_bipartite,Fred05_EAC,topchy05,wang09_pr,iam_on11_linkbased,wang11_tsmcb,franek13_pr,wu15_TKDE,wu12_tsmcb,yu15_tkde,Louren15,huang14_weac,kdd15_sec,huang15_ecfg,Huang16_neucom,Huang16_TKDE,liu2016infinite,liu17_tkde}. Ensemble clustering aims to combine multiple clusterings to obtain a probably better and more robust clustering result, which has shown advantages in finding bizarre clusters, dealing with noise, and integrating clustering solutions from multiple distributed sources \cite{wu15_TKDE}. In ensemble clustering, each input clustering is referred to as a base clustering, while the final clustering result is referred to as the consensus clustering.

In ensemble clustering, the quality of the base clusterings plays a crucial role in the consensus process. The consensus results may be badly affected by low-quality (or even ill) base clusterings. To deal with low-quality base clusterings, some efforts have been made to evaluate and weight the base clusterings to enhance the consensus performance \cite{huang14_weac,Li_WCC08,Yu14_pr}. However, these approaches \cite{huang14_weac,Li_WCC08,Yu14_pr} are developed based on an implicit assumption that all of the clusters in the same base clustering have the same reliability. They typically treat each base clustering as an individual and assign a global weight to each base clustering regardless of the diversity of the clusters inside it \cite{huang14_weac,Li_WCC08,Yu14_pr}. However, due to the noise and inherent complexity of real-world datasets, the different clusters in the same clustering may have different reliability. There is a need to respect the local diversity of ensembles and deal with the different reliability of clusters. More recently, Zhong et al. \cite{Zhong15_pr} proposed to evaluate the reliability of clusters by considering the Euclidean distances between data objects in clusters. The method in \cite{Zhong15_pr} requires access to the original data features, and its efficacy heavily relies on the data distribution of the dataset. However, in the general formulation of ensemble clustering (see Section~\ref{sec:formulation}), there is no access to the original data features. Without needing access to the data features or relying on specific assumptions about data distribution, the key problem here is how to evaluate the reliability of clusters and weight them accordingly to enhance the accuracy and robustness of the consensus clusterings.

\begin{figure}[!t]\vskip -0.07in
\begin{center}
{
{\includegraphics[width=0.69\columnwidth]{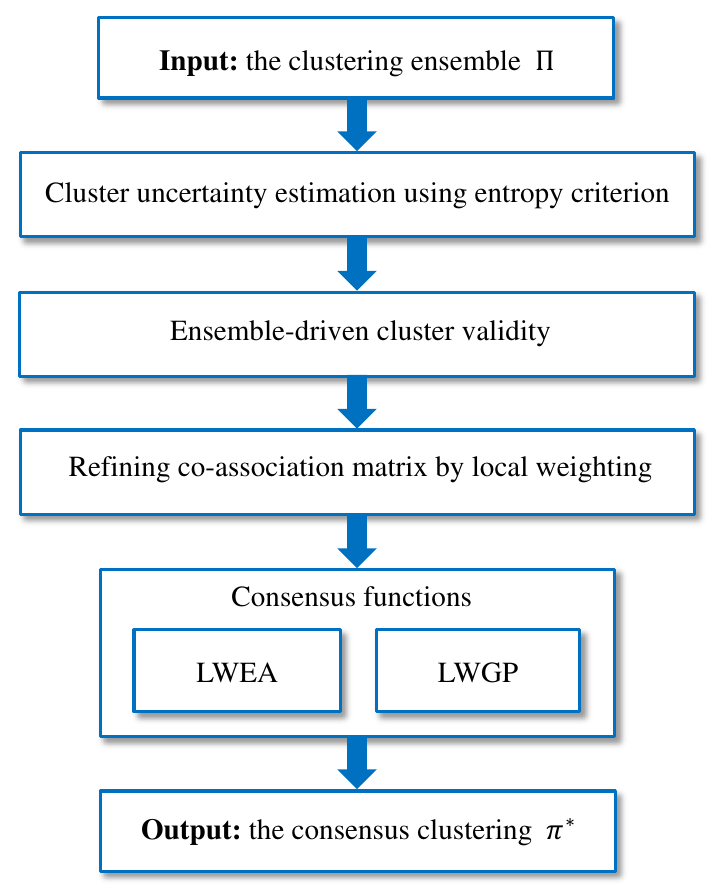}}}\vskip -0.1in
\caption{Flow diagram of the proposed approach.}\vskip -0.1in
\label{fig:flowchart}
\end{center}\vskip -0.1in
\end{figure}

Aiming to address the aforementioned problem, in this paper, we propose a novel ensemble clustering approach based on ensemble-driven cluster uncertainty estimation and local weighting strategy. The overall process of our approach is illustrated in Fig.~\ref{fig:flowchart}. We take advantage of the ensemble diversity at the cluster-level and integrate the cluster uncertainty and validity into a locally weighted scheme to enhance the consensus performance. A cluster can be viewed as a local region in the corresponding base clustering. Without needing access to the data features, in our work, the uncertainty of each cluster is estimated with regard to the cluster labels in the entire ensemble based on an entropic criterion. In particular, given a cluster, we investigate its uncertainty by considering how the objects inside this cluster are grouped in the multiple base clusterings. Based on cluster uncertainty estimation, an ensemble-driven cluster index (ECI) is then presented to measure the reliability of clusters. In this paper, we argue that the crowd of diverse clusters in the ensemble can provide an effective indication for evaluating each individual cluster. By evaluating and weighting the clusters in the ensemble via the ECI measure, we further present the concept of locally weighted co-association (LWCA) matrix, which incorporates local adaptivity into the conventional co-association (CA) matrix and serves as a summary for the ensemble of diverse clusters. Finally, to achieve the final clustering result, we propose two novel consensus functions, termed locally weighted evidence accumulation (LWEA) and locally weighted graph partitioning (LWGP), respectively, with the diversity of clusters exploited and the local weighting strategy incorporated.

For clarity, we summarize the main contributions of this paper as follows:
\begin{itemize}
  \item We propose to estimate the uncertainty of clusters by considering the distribution of all cluster labels in the ensemble using an entropic criterion, which requires no access to the original data features and makes no assumptions on the data distribution.
  \item We present an ensemble-driven cluster validity index to evaluate and weight the clusters in the ensemble, which provides an indication of reliability at the cluster-level and plays a crucial role in the local weighting scheme.
  \item We propose two novel consensus functions to construct the final clusterings based on ensemble-driven cluster uncertainty estimation and local weighting strategy.
  \item Extensive experiments have been conducted on a variety of real-world datasets, which demonstrate the superiority of the proposed ensemble clustering approach in terms of both clustering quality and efficiency.
\end{itemize}

The rest of the paper is organized as follows. The related work is reviewed in Section~\ref{sec:related_work}. The background knowledge about entropy and ensemble clustering is introduced in Section~\ref{sec:preliminary}. The proposed ensemble clustering approach based on cluster uncertainty estimation and local weighting strategy is described in Section~\ref{sec:framework}. The experimental results are reported in Section~\ref{sec:experiment}. Finally, we conclude the paper in Section~\ref{sec:conclusion}.

\section{Related Work}
\label{sec:related_work}

Ensemble learning is an important technique in machine learning, which aims to combine multiple base learners to obtain a probably better learner \cite{zhou12_ensemble_book}. Typically, there are two major directions in ensemble learning, that is, ensemble classifiers \cite{kittler98_pami,kuch04_book,Ruta05_if} and ensemble clustering \cite{vega_pons11_survey}. The ensemble classifiers technique is generally involved in supervised scenarios, while the ensemble clustering technique is generally involved in unsupervised scenarios. In this paper, our research focuses on the ensemble clustering technique, whose purpose is to combine multiple base clusterings to obtain a probably better and more robust consensus clustering. Due to its inherent unsupervised nature, ensemble clustering is still a very challenging direction in ensemble learning.

In the past decade, many ensemble clustering approaches have been developed, which can be mainly classified into three categories, i.e., the pair-wise co-occurrence based approaches \cite{Fred05_EAC,wang09_pr,iam_on11_linkbased,wang11_tsmcb,Louren15,kdd15_sec}, the graph partitioning based approaches \cite{strehl02,fern04_bipartite,huang14_weac,yu15_tkde}, and the median partition based approaches \cite{cristofor02,topchy05,franek13_pr,huang15_ecfg}.

The pair-wise co-occurrence based approaches \cite{Fred05_EAC,wang09_pr,iam_on11_linkbased,wang11_tsmcb,Louren15,kdd15_sec} typically construct a co-association (CA) matrix by considering how many times two objects occur in the same cluster among the multiple base clusterings. By exploiting the CA matrix as the similarity matrix, the conventional clustering techniques, such as the agglomerative clustering methods \cite{jain10_survey}, can be exploited to build the final clustering result. Fred and Jain \cite{Fred05_EAC} for the first time presented the concept of CA matrix and proposed the evidence accumulation clustering (EAC) method. Wang et al. \cite{wang09_pr} extended the EAC method by taking the sizes of clusters into consideration, and proposed the probability accumulation method. Iam-On et al. \cite{iam_on11_linkbased} refined the CA matrix by considering the shared neighbors between clusters to improve the consensus results. Wang \cite{wang11_tsmcb} introduced a dendrogram-like hierarchical data structure termed CA-tree to facilitate the co-association based ensemble clustering process. Louren{\c{c}}o et al. \cite{Louren15} proposed a new ensemble clustering approach which is based on the EAC paradigm and is able to determine the probabilistic assignments of data objects to clusters. Liu et al. \cite{kdd15_sec} employed spectral clustering on the CA matrix and developed an efficient ensemble clustering approach termed spectral ensemble clustering (SEC).

The graph partitioning based approaches \cite{strehl02,fern04_bipartite,huang14_weac,yu15_tkde} address the ensemble clustering problem by constructing a graph model to reflect the ensemble information. The consensus clustering is then obtained by partitioning the graph into a certain number of segments. Strehl and Ghosh \cite{strehl02} proposed three graph partitioning based ensemble clustering algorithms, i.e., cluster-based similarity partitioning algorithm (CSPA), hypergraph partitioning algorithm (HGPA), and meta-clustering algorithm (MCLA). Fern and Brodley \cite{fern04_bipartite} constructed a bipartite graph for the clustering ensemble by treating both clusters and objects as graph nodes, and obtain the consensus clustering by partitioning the bipartite graph. Yu et al. \cite{yu15_tkde} designed a double affinity propagation based ensemble clustering framework, which is able to handle the noisy attributes and obtain the final consensus clustering by the normalized cut algorithm.

The median partition based approaches \cite{cristofor02,topchy05,franek13_pr,huang15_ecfg} formulate the ensemble clustering problem into an optimization problem, which aims to find a median partition (or clustering) by maximizing the similarity between this clustering and the multiple base clusterings. The median partition problem is NP-hard \cite{topchy05}. Finding the globally optimal solution in the huge space of all possible clusterings is computationally infeasible for large datasets. Cristofor and Simovici \cite{cristofor02} proposed to obtain an approximate solution using the genetic algorithm, where clusterings are treated as chromosomes. Topchy et al. \cite{topchy05} cast the median partition problem into a maximum likelihood problem and approximately solve it by the EM algorithm. Franek and Jiang \cite{franek13_pr} cast the median partition problem into an Euclidean median problem by clustering embedding in vector spaces. Huang et al. \cite{huang15_ecfg} formulated the median partition problem into a binary linear programming problem and obtained an approximate solution by means of the factor graph theory.

These algorithms attempt to solve the ensemble clustering problem in various ways \cite{strehl02,cristofor02,fern04_bipartite,Fred05_EAC,topchy05,wang09_pr,iam_on11_linkbased,wang11_tsmcb,franek13_pr,wu15_TKDE,wu12_tsmcb,yu15_tkde,Louren15,wu15_TKDE,kdd15_sec,huang15_ecfg}. However, one common limitation to most of the existing methods is that they generally treat all clusters and all base clusterings in the ensemble equally and may suffer from low-quality clusters or low-quality base clusterings. To partially address this limitation, recently some weighted ensemble clustering approaches have been presented \cite{huang14_weac,Li_WCC08,Yu14_pr}. Li and Ding \cite{Li_WCC08} cast the ensemble clustering problem into a non-negative matrix factorization problem and proposed a weighted consensus clustering approach, where each base clustering is assigned a weight in order to improve the consensus result. Yu et al. \cite{Yu14_pr} exploited the feature selection techniques to weight and select the base clusterings. In fact, clustering selection \cite{Yu14_pr} can be viewed as a 0-1 weighting scheme, where 1 indicates selecting a clustering and 0 indicates removing a clustering. Huang et al. \cite{huang14_weac} proposed to evaluate and weight the base clusterings based on the concept of normalized crowd agreement index (NCAI), and devised two weighted consensus functions to obtain the final clustering result.

Although the above-mentioned weighted ensemble clustering approaches \cite{huang14_weac,Li_WCC08,Yu14_pr} are able to estimate the reliability of base clusterings and weight them accordingly, yet they generally treat a base clustering as a whole and neglect the local diversity of clusters inside the same base clustering. To explore the reliability of clusters, Alizadeh et al. \cite{Alizadeh14} proposed to evaluate clusters in the ensemble by averaging normalized mutual information (NMI) \cite{strehl02} between clusterings, which results in a very expensive computational cost and is not feasible for large datasets. Zhong et al. \cite{Zhong15_pr} exploited the Euclidean distances between objects to estimate the cluster reliability, which needs access to the original data features and is only applicable to numerical data. However, in the more general formulation of ensemble clustering \cite{strehl02,cristofor02,fern04_bipartite,Fred05_EAC,topchy05,wang09_pr,iam_on11_linkbased,wang11_tsmcb,franek13_pr,huang15_ecfg}, the original data features are not available in the consensus process. Moreover, by measuring the within-cluster similarity based on Euclidean distances, the efficacy of the method in \cite{Zhong15_pr} heavily relies on some implicit assumptions about data distribution, which places an unstable factor in the consensus process. Different from \cite{Zhong15_pr}, in this paper, our ensemble clustering approach requires no access to the original data features. We propose to estimate the uncertainty of clusters by considering the cluster labels in the entire ensemble based on an entropic criterion, and then present an ensemble-driven cluster index (ECI) to evaluate cluster reliability without making any assumptions on the data distribution. Further, to obtain the consensus clustering results, two novel consensus functions are developed based on cluster uncertainty estimation and local weighting strategy. Extensive experiments on a variety of real-world datasets have shown that our approach exhibits significant advantages in clustering accuracy and efficiency over the state-of-the-art approaches.

\section{Preliminaries}
\label{sec:preliminary}

\subsection{Entropy}
\label{sec:entropy}

In this section, we briefly review the concept of entropy. In information theory \cite{infoTheoBook_06}, the entropy is a measure of the uncertainty associated with a random variable. The formal definition of entropy is provided in Definition~\ref{def:entropy}.

\begin{mydef}
\label{def:entropy}
For a discrete random variable $X$, the entropy $H(X)$ is defined as
\begin{equation}
H(X)=-\sum_{x\in\mathcal{X}}p(x)\log_2 p(x),
\end{equation}
where $\mathcal{X}$ is the set of values that $X$ can take, and $p(x)$ is the probability mass function of $X$.
\end{mydef}

The joint entropy is a measure of the uncertainty associated with a set of random variables. The formal definition of joint entropy is provided in Definition~\ref{def:joint_entropy}.

\begin{mydef}
\label{def:joint_entropy}
For a pair of discrete random variables $(X, Y)$, the joint entropy $H(X,Y)$ is defined as
\begin{equation}
H(X,Y)=-\sum_{x\in\mathcal{X}}\sum_{y\in\mathcal{Y}}p(x,y)\log_2 p(x,y),
\end{equation}
where $p(x,y)$ is the joint probability of $(X,Y)$.
\end{mydef}

If and only if the two random variables $X$ and $Y$ are independent of each other, it holds that $H(X,Y)=H(X)+H(Y)$. Hence, given $n$ independent random variables $X_1, \cdots, X_{n}$, we have \cite{infoTheoBook_06}
\begin{equation}
\label{eq:entropy_N}
H(X_1,\cdots,X_n)=H(X_1)+\cdots +H(X_n).
\end{equation}

\subsection{Formulation of the Ensemble Clustering Problem}
\label{sec:formulation}

In this section, we introduce the general formulation of the ensemble clustering problem. Let $\mathcal{O}=\{o_1, \cdots, o_N\}$ be a dataset, where $o_i$ is the $i$-th data object and $N$  is the number of objects in $\mathcal{O}$. Consider $M$ partitions (or clusterings) for the dataset $\mathcal{O}$, each of which is treated as a base clustering and consists of a certain number of clusters. Formally, we denote the ensemble of $M$ base clusterings as follows:
\begin{equation}
\Pi=\{\pi^1, \cdots,\pi^M\},
\end{equation}
where
\begin{equation}
\pi^m=\{C^m_1,\cdots,C^m_{n^m}\}
\end{equation}
denotes the $m$-th base clustering in $\Pi$, $C^m_i$ denotes the $i$-th cluster in $\pi^m$, and $n^m$ denotes the number of clusters in $\pi^m$. Each cluster is a set of data objects. Obviously, the union of all clusters in the same base clustering covers the entire dataset, i.e., $\forall \pi^m\in\Pi$, $\bigcup_{i=1}^{n^m}C^m_i=\mathcal{O}$. Different clusters in the same base clustering do not intersect with each other, i.e., $\forall C^m_i, C^m_j\in\pi^m$ s.t. $i\neq j$, $C^m_i\bigcap C^m_j=\emptyset$. Let $Cls^m(o_i)$ denote the cluster in $\pi^m\in \Pi$ that object $o_i$ belongs to. That is, if $o_i$ belongs to the $k$-th cluster in $\pi^m$, i.e., $o_i\in C_k^m$, then we have $Cls^m(o_i)=C_k^m$.

For convenience, we represent the set of all clusters in the ensemble $\Pi$ as
\begin{equation}
\mathcal{C}=\{C_1,\cdots,C_{n_c}\},
\end{equation}
where $C_i$ denotes the $i$-th cluster and $n_c$ denotes the total number of clusters in $\Pi$. It is obvious that $n_c=n^1+\cdots+n^M$.

The purpose of ensemble clustering is to combine the multiple base clusterings in the ensemble $\Pi$ to obtain a probably better and more robust clustering. With regard to the difference in the input information, there are two different formulations of the ensemble clustering problem. In the first formulation, the ensemble clustering system only takes the multiple base clusterings as input and has no access to the original data features \cite{strehl02,cristofor02,fern04_bipartite,Fred05_EAC,topchy05,wang09_pr,iam_on11_linkbased,wang11_tsmcb,franek13_pr,huang15_ecfg,yi_icdm12}. In the other formulation, the ensemble clustering system takes both the multiple base clusterings and the original data features as inputs \cite{Zhong15_pr,vega_pons10}. In this paper, we comply with the first formulation of the ensemble clustering problem, which is also the common practice for most of the existing ensemble clustering approaches \cite{vega_pons11_survey}. Hence, in our formulation, the input is the clustering ensemble $\Pi$, and the output is the consensus clustering $\pi^*$.

\section{Locally Weighted Ensemble Clustering}
\label{sec:framework}

In this paper, we propose a novel ensemble clustering approach based on ensemble-driven cluster uncertainty estimation and local weighting strategy. Without needing access to the original data features or making some assumptions about data distribution, we exploit the ensemble information to estimate the uncertainty (or unreliability) of clusters based on an entropic criterion. With the cluster uncertainty obtained, an ensemble-driven cluster validity index termed ECI is presented to evaluate the reliability of each cluster with the help of the cluster labels in the clustering ensemble. In this paper, we argue that the diverse clusters in the ensemble can provide an effective indication for evaluating the reliability of each individual cluster. Then, we refine the conventional CA matrix using a local weighting strategy based on the ECI measure, and introduce the concept of locally weighted co-association (LWCA) matrix , which serves as a summary for the ensemble with diverse clusters. To obtain the final clustering results, in this paper, two novel consensus functions are further presented, that is, LWEA and LWGP. In the following of this section, we will describe each step of our approach in detail.

\subsection{Measuring Cluster Uncertainty in Ensembles}
\label{sec:measure_uncertainty}

In the general formulation of ensemble clustering, there is no access to the original data features. To evaluate the reliability of each cluster, we appeal to the concept of entropy with the help of the cluster labels in the entire ensemble.

As introduced in Section~\ref{sec:entropy}, entropy is a measure of uncertainty associated with a random variable. Each cluster is a set of data objects. Given a cluster $C_i\in\mathcal{C}$ and a base clustering $\pi^m\in\Pi$, if cluster $C_i$ does not belong to $\pi^m$, then it is possible that the objects in $C_i$ belong to more than one cluster in $\pi^m$. In fact, the objects in $C_i$ may belong to at most $n^m$ different clusters in $\pi^m$, where $n^m$ is the total number of clusters in $\pi^m$. The uncertainty (or entropy) of $C_i$ w.r.t. $\pi^m$ can be computed by considering how the objects in $C_i$ are clustered in $\pi^m$.

\begin{mydef}
\label{def:uncertain_pi_m}
Given the ensemble $\Pi$, the uncertainty of cluster $C_i$ w.r.t. the base clustering $\pi^m\in\Pi$ is computed as
\begin{equation}
H^m(C_i)=-\sum_{j=1}^{n^m}p(C_i,C_j^m)\log_2 p(C_i,C_j^m)
\end{equation}
with
\begin{equation}
p(C_i,C_j^m)=\frac{|C_i\bigcap C_j^m|}{|C_i|},
\end{equation}
where $n^m$ is the number of clusters in $\pi^m$, $C_j^m$ is the $j$-th cluster in $\pi^m$, $\bigcap$ computes the intersection of two sets (or clusters), and $|C_i|$ outputs the number of objects in $C_i$.
\end{mydef}

The formal definition of the cluster uncertainty w.r.t. a base clustering is given in Definition~\ref{def:uncertain_pi_m}. Because it holds that $p(C_i,C_j^m)\in[0,1]$ for any $i$, $j$, $m$, so we have $H^m(C_i)\in[0,+\infty)$. When all the objects in $C_i$ belong to the same cluster in $\pi^m$, the uncertainty of $C_i$ w.r.t. $\pi^m$ reaches its minimum, i.e., zero. When the objects in $C_i$ belong to more different clusters in $\pi^m$, the uncertainty of $C_i$ w.r.t. $\pi^m$ typically gets greater, which indicates that the objects in $C_i$ are less likely to be in the same cluster with regard to $\pi^m$.

Without loss of generality, based on the assumption that the base clusterings in the ensemble are independent \cite{vega_pons11_survey}, the uncertainty (or entropy) of $C_i$ w.r.t. the entire ensemble $\Pi$ can be computed by summing up the uncertainty of $C_i$ w.r.t. the $M$ base clusterings in $\Pi$ according to Eq.~(\ref{eq:entropy_N}). Its formal definition is given in Definition~\ref{def:uncertain_Pi}.

\begin{mydef}
\label{def:uncertain_Pi}
Given the ensemble $\Pi$, the uncertainty of cluster $C_i$ w.r.t. the entire ensemble $\Pi$ is computed as
\begin{equation}
H^{\Pi}(C_i)=\sum_{m=1}^{M}H^m(C_i),
\end{equation}
where $M$ is the number of base clusterings in $\Pi$.
\end{mydef}

Intuitively, the uncertainty of $C_i$ w.r.t. $\Pi$ reflects how the objects in $C_i$ are clustered in the ensemble of multiple base clusterings. If the objects in $C_i$ belong to the same cluster in each of the base clusterings, which can be viewed as that all base clusterings \emph{agree} that  the objects in $C_i$ should be assigned to the same cluster, then the uncertainty of $C_i$ w.r.t. $\Pi$ reaches its minimum, i.e., zero. When the uncertainty of $C_i$ w.r.t. $\Pi$ gets larger, it is indicated that the objects in $C_i$ are less likely to be in the same cluster with consideration to the ensemble of multiple base clusterings.

\begin{figure}[!t]\vskip -0.22 in
\begin{center}
{\subfigure[$\pi^1$]
{\includegraphics[width=0.32\linewidth]{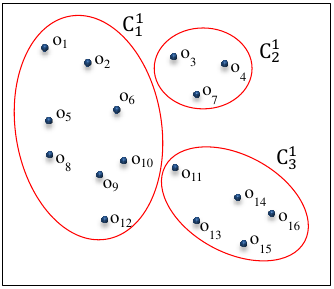}\label{fig:entropyDemo1}}}
{\subfigure[$\pi^2$]
{\includegraphics[width=0.32\linewidth]{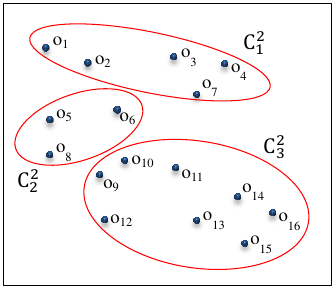}\label{fig:entropyDemo2}}}
{\subfigure[$\pi^3$]
{\includegraphics[width=0.32\linewidth]{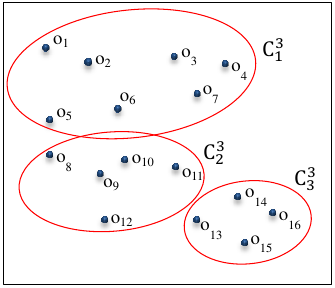}\label{fig:entropyDemo3}}}\vskip -0.11 in
\caption{Illustration of an ensemble of three base clusterings, namely, $\pi^1$, $\pi^2$, and $\pi^3$.}
\label{fig:entropyDemos}
\end{center}\vskip -0.15 in
\end{figure}

\begin{table}[!t] \vskip -0.05 in
\centering
\caption{Computation of cluster uncertainty and ECI (with $\theta=0.5$) for the clusters in the ensemble shown in Fig.~\ref{fig:entropyDemos}.}\vskip -0.1in
\label{table:entropyDemos}
\begin{tabular}{m{1.2cm}<{\centering}|m{0.9cm}<{\centering}|m{2.6cm}<{\centering}m{2.3cm}<{\centering}}
\toprule
Base Clustering		&Cluster	&Cluster Uncertainty w.r.t. the Ensemble &ECI\\
\midrule
\multirow{3}{*}{$\pi^1$}	&$C_1^1$&$H^{\Pi}(C_1^1)=2.56$	&$ECI(C_1^1)=0.18$\\
             &$C_2^1$          &$H^{\Pi}(C_2^1)=0.00$	&$ECI(C_2^1)=1.00$\\
             &$C_3^1$	&$H^{\Pi}(C_3^1)=0.72$	&$ECI(C_3^1)=0.62$\\
\midrule
\multirow{3}{*}{$\pi^2$}	&$C_1^2$&$H^{\Pi}(C_1^2)=0.97$	&$ECI(C_1^2)=0.52$\\
             &$C_2^2$          &$H^{\Pi}(C_2^2)=0.92$	&$ECI(C_2^2)=0.54$\\
             &$C_3^2$	&$H^{\Pi}(C_3^2)=1.95$	&$ECI(C_3^2)=0.27$\\
\midrule
\multirow{3}{*}{$\pi^3$}	&$C_1^3$&$H^{\Pi}(C_1^3)=1.85$	&$ECI(C_1^3)=0.29$\\
             &$C_2^3$          &$H^{\Pi}(C_2^3)=1.44$	&$ECI(C_2^3)=0.38$\\
             &$C_3^3$	&$H^{\Pi}(C_3^3)=0.00$	&$ECI(C_3^3)=1.00$\\
\bottomrule
\end{tabular}\vskip -0.15in
\end{table}

We provide an example\footnote{We have fixed an error in the computation of this example which happened in the previous version. We would like to thank Mr. Recep Do\v{g}a Siyli for pointing out this issue.} in Fig.~\ref{fig:entropyDemos} and Table~\ref{table:entropyDemos} to show the computation of cluster uncertainty w.r.t. an ensemble of three base clusterings. For the dataset $\mathcal{O}=\{o_1,\cdots,o_{16}\}$ with 16 data objects, three base clusterings ($\pi^1$, $\pi^2$, and $\pi^3$) are generated, each of which consists of three clusters (as illustrated in Fig.~\ref{fig:entropyDemos}). Of the three clusters in $\pi^1$, $C_1^1$ contains eight objects, $C_2^1$ contains three objects, and $C_3^1$ contains five objects. Then, we proceed to compute the uncertainty of the three clusters in $\pi^1$ w.r.t. the ensemble. The eight objects in cluster $C_1^1$ belong to three different clusters in $\pi^2$. According to Definition~\ref{def:uncertain_pi_m}, with $p(C_1^1,C_1^2)=\frac{2}{8}$, $p(C_1^1,C_2^2)=\frac{3}{8}$, and $p(C_1^1,C_3^2)=\frac{3}{8}$, the uncertainty of $C_1^1$ w.r.t. base clustering $\pi^2$ is computed as $H^2(C_1^1) =-\frac{2}{8}\cdot\log_2 \frac{2}{8}-\frac{3}{8}\cdot\log_2 \frac{3}{8}-\frac{3}{8}\cdot\log_2 \frac{3}{8}\approx 1.56$. Similarly, we can obtain $H^3(C_1^1)=1$. It is obvious that the uncertainty of cluster $C_1^1$ w.r.t. the base clustering that contains it equals zero, i.e, $H^1(C_1^1)=0$. Therefore, the uncertainty of cluster $C_1^1$ w.r.t. the entire ensemble $\Pi$ can be computed as $H^{\Pi}(C_1^1)=0+1.56+1=2.56$. In a similar way, the uncertainty of the other clusters in $\Pi$ can be obtained  (see Table~\ref{table:entropyDemos}). It is noteworthy that the three objects in $C_2^1$ belong to the same cluster in each of the three base clusterings in $\Pi$, i.e., all base clusterings in $\Pi$ agree that the objects in $C_2^1$ should be in the same cluster. Thereby the uncertainty of $C_2^1$ w.r.t. $\Pi$ reaches the minimum value, that is, $H^{\Pi}(C_2^1)=0$. As shown in Table~\ref{table:entropyDemos}, of the nine clusters in $\Pi$, $C_1^1$ is the cluster with the greatest uncertainty, while $C_2^1$ and $C_3^3$ are the two most stable clusters. For clarity, in the following, when we refer to cluster uncertainty without mentioning whether it is with respect to a base clustering or with respect to the ensemble, we mean cluster uncertainty w.r.t. the ensemble.

\subsection{Ensemble-Driven Cluster Validity}
\label{sec:ECI}
Having obtained the uncertainty (or entropy) of each cluster in the clustering ensemble, we further propose an ensemble-driven cluster index (ECI), which measures the reliability of clusters by considering their uncertainty w.r.t. the ensemble.

\begin{mydef}
\label{def:ECI}
Given an ensemble $\Pi$ with $M$ base clusterings, the ensemble-driven cluster index (ECI) for a cluster $C_i$  is defined as
\begin{equation}
\label{eq:ECI}
ECI(C_i)=\mathrm{e}^{-\frac{H^{\Pi}(C_i)}{\theta\cdot M}},
\end{equation}
where $\theta>0$ is a parameter to adjust the influence of the cluster uncertainty over the index.
\end{mydef}

\begin{figure}[!t]\vskip -0.22in
\begin{center}
{
{\includegraphics[width=0.55\columnwidth]{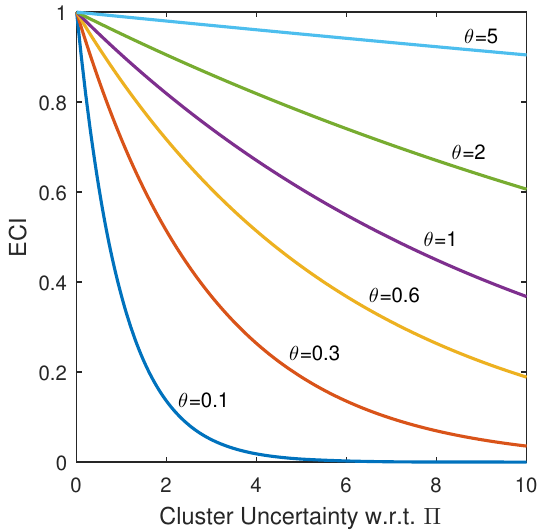}}}
\caption{Correlation between ECI and cluster uncertainty with different parameters $\theta$.}\vskip -0.05in
\label{fig:uncertainty_vs_eci}
\end{center}\vskip -0.1in
\end{figure}

The formal definition of ECI is given in Definition~\ref{def:ECI}. According to the definition, because $H^{\Pi}(C_i)\in [0,+\infty)$, it holds that $ECI(C_i)\in (0,1]$ for any $C_i \in \mathcal{C}$.
Obviously, smaller uncertainty of a cluster leads to a greater ECI value. As an example, Table~\ref{table:entropyDemos} shows the ECI values for the clusters in the ensemble illustrated in Fig.~\ref{fig:entropyDemos}.

When the uncertainty of a cluster $C_i$ reaches its minimum, i.e., $H^{\Pi}(C_i)=0$, its ECI will thereby reaches its maximum, i.e., $ECI(C_i)=1$. The ECI of a cluster approaches zero when its cluster uncertainty approaches infinity. A parameter $\theta$ is adopted in the computation of ECI to adjust the influence of the cluster uncertainty over the index (see Eq.~(\ref{eq:ECI})). As shown in Fig.~\ref{fig:uncertainty_vs_eci}, when setting $\theta$ to small values, e.g., setting $\theta<0.1$, the ECI decreases dramatically as the cluster uncertainty increases. When setting $\theta$ to large values, the difference between the ECI values of high-uncertainty clusters and low-uncertainty ones will be narrowed down. Empirically, it is suggested that the parameter $\theta$ be set in the interval of $[0.2,1]$. The consensus performance of our approach with different parameters $\theta$ is evaluated by extensive experiments. Please see Section~\ref{sec:para_anal} for more details.

\subsection{Refining Co-association Matrix by Local Weighting}
\label{sec:LWCA}

The co-association (CA) matrix is first proposed by Fred and Jain \cite{Fred05_EAC}, which reflects how many times two data objects are grouped into the same cluster among the multiple base clusterings in the ensemble.
\begin{mydef}
\label{def:CA}
Given an ensemble $\Pi$, the co-association (CA) matrix is computed as
\begin{equation}
A=\{a_{ij}\}_{N\times N}
\end{equation}
with
\begin{align}
a_{ij} =& \frac{1}{M}\cdot\sum_{m=1}^M \delta^m_{ij},\\
\delta^m_{ij}=&\begin{cases}1,&\text{if~}Cls^m(o_i)=Cls^m(o_j),\\
0,&\text{otherwise,}
\end{cases}
\end{align}
where $Cls^m(o_i)$ denotes the cluster in $\pi^m\in \Pi$ that object $o_i$ belongs to.
\end{mydef}

The CA matrix is a classical and widely used tool for dealing with the ensemble clustering problem \cite{Fred05_EAC,wang11_tsmcb,yu15_tkde,Mimaroglu11_pr}. Despite the significant success, one limitation of the CA matrix is that it treats all clusters and all base clusterings in the ensemble equally and lack the ability to evaluate and weight the ensemble members w.r.t. their reliability. Huang et al. \cite{huang14_weac} exploited the NCAI index to weight the base clusterings and thereby construct a weighted co-association (WCA) matrix, which, however, only considers the reliability of base clusterings, but still neglects the cluster-wise diversity inside the same base clustering.

Different from the (globally) weighting strategy \cite{huang14_weac} that treats each base clustering as a whole, in this section, we refine the CA matrix by a local weighting strategy based on the ensemble-driven cluster validity and propose the concept of locally weighted co-association (LWCA) matrix.

\begin{mydef}
\label{def:LWCA}
Given an ensemble $\Pi$, the locally weighted co-association (LWCA) matrix is computed as
\begin{equation}
\label{eq:lwca_A}
\tilde{A}=\{\tilde{a}_{ij}\}_{N\times N}
\end{equation}
with
\begin{align}
\tilde{a}_{ij} =& \frac{1}{M}\cdot\sum_{m=1}^M w_i^m\cdot\delta^m_{ij},\label{eq:lwca_aij}\\
w_i^m =& ECI\left(Cls^m(o_i)\right),\label{eq:lwca_wim}\\
\delta^m_{ij}=&\begin{cases}1,&\text{if~}Cls^m(o_i)=Cls^m(o_j),\label{eq:lwca_delta2}\\
0,&\text{otherwise,}
\end{cases}
\end{align}
where $Cls^m(o_i)$ denotes the cluster in $\pi^m\in \Pi$ that object $o_i$ belongs to.
\end{mydef}

A cluster can be viewed as a local region in a base clustering. To take into consideration the different reliability of clusters in the ensemble, the weighting term $w_i^m$ is incorporated to assign weights to clusters via the ECI measure (see Definition~\ref{def:LWCA}). The intuition is that the objects that co-occur in more reliable clusters (with higher ECI values) are more likely to belong to the same cluster in the true clustering. With the local weighting strategy, the LWCA matrix not just considers how many times two objects occur in the same cluster among the multiple base clusterings, but also reflects how reliable the clusters in the ensemble are.

\subsection{Consensus Functions}

In this paper, based on ensemble-driven cluster uncertainty estimation and local weighting strategy, we further propose two novel consensus functions, i.e., locally weighted evidence accumulation (LWEA) and locally weighted graph partitioning (LWGP), which will be described in Section~\ref{sec:LWEA} and Section~\ref{sec:LWGP}, respectively.

\subsubsection{Locally Weighted Evidence Accumulation (LWEA)}
\label{sec:LWEA}

In this section, we introduce the consensus function termed LWEA, which is based on hierarchical agglomerative clustering.

Hierarchical agglomerative clustering is a widely used clustering technique \cite{jain10_survey}, which typically takes a similarity matrix as input and performs region merging iteratively to achieve a dendrogram, i.e., a hierarchical representation of clusterings. Here, we exploit the LWCA matrix (see Definition~\ref{def:LWCA}) as the initial similarity matrix, denoted as
\begin{equation}
S^{(0)}=\{S^{(0)}_{ij}\}_{N\times N},
\end{equation}
with
\begin{equation}
S^{(0)}_{ij}=\tilde{a}_{ij},
\end{equation}
where $\tilde{a}_{ij}$ is the $(i,j)$-th entry in the LWCA matrix. The $N$ original data objects are treated as the $N$ initial regions. Formally, we denote the set of initial regions as follows:
\begin{equation}
\mathcal{R}^{(0)}=\{R^{(0)}_1,\cdots,R^{(0)}_N\},
\end{equation}
where
\begin{align}
R^{(0)}_i = \{o_i\}, \text{~~for~}i=1,\cdots,N.
\end{align}
denotes the $i$-th region in $\mathcal{R}^{(0)}$. Note that each initial region contains exactly one data object.

With the initial similarity matrix and the initial regions constructed, the region merging process is then performed iteratively. In each step of region merging, the two regions with the highest similarity will be merged into a new and larger region and thereby the set of regions will be updated. The set of the obtained regions in the $t$-th step is denoted as follows:
\begin{equation}
\mathcal{R}^{(t)}=\{R^{(t)}_1,\cdots,R^{(t)}_{|\mathcal{R}^{(t)}|}\},
\end{equation}
where $R^{(t)}_i$ denotes the $i$-th region and $|\mathcal{R}^{(t)}|$ denotes the number of regions in $\mathcal{R}^{(t)}$. After each step of region merging, to get prepared for the next iteration, the similarity matrix will be updated according to the new set of regions. Typically, we adopt the average-link (AL), which is a classical agglomerative clustering method \cite{jain10_survey}, to update the similarity matrix for the $t$-step. That is
\begin{equation}
S^{(t)}=\{S^{(t)}_{ij}\}_{|\mathcal{R}^{(t)}|\times |\mathcal{R}^{(t)}|}
\end{equation}
with
\begin{equation}
S^{(t)}_{ij}=\frac{1}{|R^{(t)}_i|\cdot|R^{(t)}_j|}\sum_{o_k\in R^{(t)}_i,o_l\in R^{(t)}_j}\tilde{a}_{kl},
\end{equation}
where $|R^{(t)}_i|$ denotes the number of objects in the region $R^{(t)}_i$.

By iterative region merging, a dendrogram is constructed. The root of the dendrogram is the entire dataset, while the leaves of it are the original data objects. Each level of the dendrogram represents a clustering with a certain number of clusters. Therefore, the final clustering result can be obtained by specifying a number of clusters for the dendrogram.

For clarity, the overall algorithm of LWEA is summarized in Algorithm 1.

\begin{figure}[!h]
\textbf{Algorithm 1 (Locally Weighted Evidence Accumulation)}\\
\small{ {\bfseries Input:} $\Pi$, $k$.
\begin{algorithmic}[1]
    \STATE Compute the uncertainty of the clusters in $\Pi$ w.r.t. Definition~\ref{def:uncertain_Pi}.\\
    \STATE Compute the ECI measures of the clusters in $\Pi$ w.r.t. Definition~\ref{def:ECI}.\\
    \STATE Construct the LWCA matrix w.r.t. Definition~\ref{def:LWCA}.\\
    \STATE Initialize the set of regions $\mathcal{R}^{(0)}$ and the similarity matrix $S^{(0)}$.
    \STATE Construct the dendrogram iteratively:\\
    \textbf{for} {$t=1,2,\cdots,\tilde{N}-1$}\\
     ~~~~Merge the two most similar regions in $\mathcal{R}^{(t-1)}$ w.r.t. $S^{(t-1)}$.\\
     ~~~~Obtain the new set of regions $\mathcal{R}^{(t)}$.\\
     ~~~~Obtain the new similarity matrix $S^{(t)}$.\\
    \textbf{end for}\\
    \STATE Obtain the clustering with $k$ clusters in the dendrogram.\\
\end{algorithmic}
{\bfseries Output:} the consensus clustering $\pi^*$.}
\end{figure}

\subsubsection{Locally Weighted Graph Partitioning (LWGP)}
\label{sec:LWGP}

In this section, we introduce the consensus function termed LWGP, which is based on bipartite graph formulating and partitioning.

\begin{figure}[!t]
\begin{center}
{
{\includegraphics[width=0.65\linewidth]{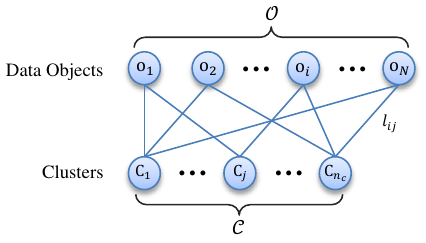}}\hskip 0.15in}
\caption{Illustration of the locally weighted bipartite graph (LWBG).}
\label{fig:LWBG}
\end{center}\vskip -0.1in
\end{figure}

To construct the bipartite graph, we treat both clusters and objects as  graph nodes. A link between two nodes exists if and only if one node is a data object and the other node is the cluster that contains it (see Fig.~\ref{fig:LWBG}). Given an object $o_i\in\mathcal{O}$  and a cluster $C_j\in\mathcal{C}$ such that $o_i\in C_j$, the link weight between them is decided by the ECI value of $C_j$, i.e., the weight of a link is correlated to the reliability of the cluster that it connects to. Hence, with the ECI measure incorporated, the bipartite graph not only considers the belong-to relationship between objects and clusters, but also reflects the local reliability, i.e., the reliability of clusters, in the ensemble. Formally, the locally weighted bipartite graph (LWBG) is defined in Definition~\ref{def:LWBG}.

\begin{mydef}
\label{def:LWBG}
The locally weighted bipartite graph (LWBP) is defined as
\begin{equation}
G=(\mathcal{V},\mathcal{L}),
\end{equation}
where $\mathcal{V}=\mathcal{O}\bigcup\mathcal{C}$ is the node set and $\mathcal{L}$ is the link set. The link weight between two nodes $v_i$ and $v_j$ is defined as
\begin{equation}
l_{ij}=\begin{cases}ECI(v_j), &\text{if~}v_i\in\mathcal{O},v_j\in\mathcal{C},\text{and }v_i\in v_j,\\
ECI(v_i), &\text{if~}v_j\in\mathcal{O},v_i\in\mathcal{C},\text{and }v_j\in v_i,\\
0, &\text{otherwise.}\\
\end{cases}
\end{equation}
\end{mydef}

Having constructed the LWBG according to Definition~\ref{def:LWBG}, we proceed to partition the graph using the Tcut algorithm \cite{CVPR12_Li}, which is able to take advantage of the bipartite graph structure to greatly facilitate the computation of the graph partitioning process. The graph is partitioned into a certain number of disjoint node sets. The object nodes in the same segment are treated as a cluster, and hence the final clustering result can be obtained.

For clarity, we summarize the LWGP algorithm in Algorithm 2.

\begin{figure}[!htb]
\textbf{Algorithm 2 (Locally Weighted Graph Partitioning)}\\
\small{ {\bfseries Input:} $\Pi$, $k$.
\begin{algorithmic}[1]
    \STATE Compute the uncertainty of the clusters in $\Pi$ w.r.t. Definition~\ref{def:uncertain_Pi}.\\
    \STATE Compute the ECI measures of the clusters in $\Pi$ w.r.t. Definition~\ref{def:ECI}.\\
    \STATE Build the LWBG graph w.r.t. Definition~\ref{def:LWBG}.\\
    \STATE Partition the LWBG into a certain number of segments using the Tcut algorithm \cite{CVPR12_Li}.\\
    \STATE Treat objects in the same segment as a cluster and form clusters for the entire dataset.
    \STATE Obtain the consensus clustering by the obtained clusters .\\
\end{algorithmic}
{\bfseries Output:} the consensus clustering $\pi^*$.}
\end{figure}

\section{Experiments}
\label{sec:experiment}

In this section, we conduct experiments on a variety of real-world datasets to compare the proposed methods against the state-of-the-art ensemble clustering methods. The MATLAB source code and experimental data of this work are available at: https://www.researchgate.net/publication/316681928.

\subsection{Datasets and Evaluation Methods}
\label{sec:dataset_and_eval}

In our experiments, fifteen real-world datasets are used, namely, \emph{Caltech20}, \emph{Forest Covertype (FCT)}, \emph{Image Segmentation (IS)}, \emph{ISOLET}, \emph{Letter Recognition (LR)}, \emph{Landsat Satellite (LS)}, \emph{Multiple Features (MF)}, \emph{MNIST}, \emph{Optical Digit Recognition (ODR)}, \emph{Pen Digit (PD)}, \emph{Semeion}, \emph{Steel Plates Faults (SPF)}, \emph{Texture}, \emph{Vehicle Silhouettes (VS)}, and \emph{USPS}. Following the practice of \cite{meap13}, we select 20 representative categories out of the 101 categories in the \emph{Caltech} dataset\footnote{http://www.vision.caltech.edu/feifeili/Datasets.htm} to form the \emph{Caltech20} dataset. The \emph{MNIST} and \emph{USPS} datasets are from Dr. Sam Roweis's website\footnote{http://www.cs.nyu.edu/\%7eroweis/data.html}, where a subset of $5,000$ objects is used here for the \emph{MNIST} dataset. The other twelve datasets are from the UCI machine learning repository\footnote{\url{http://archive.ics.uci.edu/ml}}. The details of the fifteen datasets are given in Table~\ref{table:datasets}.

\begin{table}[!t]%
\centering
\caption{Description of the benchmark datasets.}\vskip -0.1 in
\label{table:datasets}
\begin{center}
\begin{tabular}{p{1.6cm}<{\centering}|p{1.5cm}<{\centering}p{1.3cm}<{\centering}p{1.3cm}<{\centering}}
\toprule
Dataset         &\#Object     &\#Attribute      &\#Class\\
\midrule
\emph{Caltech20}      &2,386      &30,000    &20\\
\emph{FCT}                &3,780  &54      &7\\
\emph{IS}                &2,310  &19     &7\\
\emph{ISOLET}           &7,797   &617    &26\\
\emph{LR}           &20,000     &16     &26\\
\emph{LS}                &6,435  &36      &6\\
\emph{MF}                &2,000  &649      &10\\
\emph{MNIST}                &5,000  &784      &10\\
\emph{ODR}                &5,620  &64      &10\\
\emph{PD}                &10,992  &16      &10\\
\emph{Semeion}      &1,593      &256    &10\\
\emph{SPF}      &1,941      &27    &7\\
\emph{Texture}      &5,500      &40    &11\\
\emph{VS}                 &846    &18     &4\\
\emph{USPS}                 &11,000    &256     &10\\
\bottomrule
\end{tabular}
\end{center}
\end{table}

Two widely used evaluation measures, i.e., normalized mutual information (NMI) \cite{strehl02} and adjusted rand index (ARI) \cite{vinh2010_ARI}, are used to evaluate the quality of clusterings. Note that larger values of NMI and ARI indicate better clustering results.

The NMI measure provides a sound indication of the shared information between two clusterings. Let $\pi'$ be the test clustering and $\pi^G$ the ground-truth clustering. The NMI score of $\pi'$ w.r.t. $\pi^G$ is defined as follows \cite{strehl02}:
\begin{equation}
\label{eq:nmi}
NMI(\pi', \pi^G)=\frac{\sum_{i=1}^{n'}\sum_{j=1}^{n^G}n_{ij}\log\frac{n_{ij}n}{n_i'n_j^G}}{\sqrt{\sum_{i=1}^{n'}n_i'\log\frac{n_i'}{n}\sum_{j=1}^{n^G}n_j^G\log\frac{n_j^G}{n}}},
\end{equation}
where $n'$ is the number of clusters in $\pi'$, $n^G$ is the number of clusters in $\pi^G$, $n_i'$ is the number of objects in the $i$-th cluster of $\pi'$, $n_j^G$ is the number of objects in the $j$-th cluster of $\pi^G$, and $n_{ij}$ is the number of common objects shared by cluster $i$ in $\pi'$ and cluster $j$ in $\pi^G$.

The ARI is a generalization of the rand index (RI) \cite{rand71}, which is computed by considering the number of pairs of objects on which two clusterings agree or disagree. Specifically, the ARI score of $\pi'$ w.r.t. $\pi^G$ is computed as follows \cite{vinh2010_ARI}:
\begin{align}
&ARI(\pi', \pi^G)=\nonumber\\
& \frac{2(N_{00}N_{11}-N_{01}N_{10})}{(N_{00}+N_{01})(N_{01}+N_{11})+(N_{00}+N_{10})(N_{10}+N_{11})},
\end{align}
where $N_{11}$ is the number of object pairs that appear in the same cluster in both $\pi'$ and $\pi^G$, $N_{00}$ is the number of object pairs that appear in different clusters in $\pi'$ and $\pi^G$, $N_{10}$ is the number of object pairs that appear in the same cluster in  $\pi'$ but in different clusters in $\pi^G$, and $N_{01}$ is the number of object pairs that appear in different clusters in  $\pi'$ but in the same cluster in $\pi^G$.

\begin{table}[!t]
\centering
\caption{The performance of LWEA with varying parameters $\theta$ (in terms of NMI).}\vskip -0.1 in
\label{table:para_weac}
\begin{tabular}{m{1cm}<{\centering}m{0.50cm}<{\centering}m{0.50cm}<{\centering}m{0.50cm}<{\centering}m{0.50cm}<{\centering}m{0.50cm}<{\centering}m{0.50cm}<{\centering}m{0.50cm}<{\centering}m{0.50cm}<{\centering}}
\toprule
\multirow{2}{*}{Dataset}             &\multicolumn{8}{c}{$\theta$}\\
\cmidrule{2-9}
             &0.1          &0.2          &0.4          &0.6          &0.8         &1&2&4\\
\midrule
\emph{Caltech20} 	&$0.416$	&$0.473$	&$0.477$	&$0.472$	&$0.467$	&$0.465$	&$0.460$	&$0.458$\\
\emph{FCT}	&$0.230$ 	&$0.244$ 	&$0.243$ 	&$0.243$ 	&$0.238$ 	&$0.237$ 	&$0.232$ 	&$0.229$\\
\emph{IS}	&$0.676$ 	&$0.670$ 	&$0.640$ 	&$0.626$ 	&$0.621$ 	&$0.619$ 	&$0.615$ 	&$0.615$\\
\emph{ISOLET}	&$0.624$ 	&$0.753$ 	&$0.754$ 	&$0.751$ 	&$0.749$ 	&$0.748$ 	&$0.747$ 	&$0.747$\\
\emph{LR}	&$0.108$ 	&$0.441$ 	&$0.449$ 	&$0.445$ 	&$0.442$ 	&$0.441$ 	&$0.438$ 	&$0.437$\\
\emph{LS}	&$0.574$ 	&$0.605$ 	&$0.632$ 	&$0.628$ 	&$0.623$ 	&$0.621$ 	&$0.608$ 	&$0.604$\\
\emph{MF}	&$0.667$ 	&$0.686$ 	&$0.681$ 	&$0.670$ 	&$0.668$ 	&$0.663$ 	&$0.655$ 	&$0.647$\\
\emph{MNIST}	&$0.461$ 	&$0.636$ 	&$0.655$ 	&$0.649$ 	&$0.638$ 	&$0.635$ 	&$0.615$ 	&$0.608$\\
\emph{ODR}	&$0.795$ 	&$0.839$ 	&$0.835$ 	&$0.835$ 	&$0.835$ 	&$0.830$ 	&$0.825$ 	&$0.817$\\
\emph{PD}	&$0.781$	&$0.801$	&$0.794$	&$0.784$	&$0.778$	&$0.775$	&$0.762$	&$0.756$\\
\emph{Semeion}	&$0.549$ 	&$0.651$ 	&$0.663$ 	&$0.658$ 	&$0.658$ 	&$0.657$ 	&$0.650$ 	&$0.645$\\
\emph{SPF}	&$0.169$	&$0.170$	&$0.163$	&$0.160$	&$0.155$	&$0.153$	&$0.152$	&$0.152$\\
\emph{Texture}	&$0.767$	&$0.796$	&$0.784$	&$0.769$	&$0.759$	&$0.753$	&$0.738$	&$0.729$\\
\emph{VS}	&$0.156$	&$0.157$	&$0.160$	&$0.162$	&$0.164$	&$0.162$	&$0.162$	&$0.160$\\
\emph{USPS}	&$0.534$ 	&$0.659$ 	&$0.660$ 	&$0.641$ 	&$0.628$ 	&$0.625$ 	&$0.602$ 	&$0.597$\\
\bottomrule
\end{tabular}
\end{table}

\begin{table}[!t]
\centering
\caption{The performance of LWGP with varying parameters $\theta$ (in terms of NMI).}\vskip -0.1 in
\label{table:para_LWGP}
\begin{tabular}{m{1cm}<{\centering}m{0.50cm}<{\centering}m{0.50cm}<{\centering}m{0.50cm}<{\centering}m{0.50cm}<{\centering}m{0.50cm}<{\centering}m{0.50cm}<{\centering}m{0.50cm}<{\centering}m{0.50cm}<{\centering}}
\toprule
\multirow{2}{*}{Dataset}             &\multicolumn{8}{c}{$\theta$}\\
\cmidrule{2-9}
             &0.1          &0.2          &0.4          &0.6          &0.8         &1&2&4 \\
\midrule
\emph{Caltech20} 	&$0.415$	&$0.477$	&$0.459$	&$0.457$	&$0.458$	&$0.457$	&$0.455$	&$0.454$\\
\emph{FCT}	&$0.239$ 	&$0.232$ 	&$0.242$ 	&$0.244$ 	&$0.243$ 	&$0.242$ 	&$0.239$ 	&$0.237$\\
\emph{IS}	&$0.670$ 	&$0.682$ 	&$0.658$ 	&$0.652$ 	&$0.646$ 	&$0.639$ 	&$0.632$ 	&$0.628$\\
\emph{ISOLET}	&$0.727$ 	&$0.750$ 	&$0.744$ 	&$0.743$ 	&$0.743$ 	&$0.742$ 	&$0.741$ 	&$0.741$\\
\emph{LR}	&$0.327$ 	&$0.453$ 	&$0.447$ 	&$0.444$ 	&$0.444$ 	&$0.443$ 	&$0.442$ 	&$0.441$\\
\emph{LS}	&$0.562$ 	&$0.618$ 	&$0.650$ 	&$0.647$ 	&$0.644$ 	&$0.638$ 	&$0.632$ 	&$0.626$\\
\emph{MF}	&$0.635$ 	&$0.685$ 	&$0.696$ 	&$0.687$ 	&$0.684$ 	&$0.678$ 	&$0.671$ 	&$0.665$\\
\emph{MNIST}	&$0.375$ 	&$0.624$ 	&$0.644$ 	&$0.645$ 	&$0.643$ 	&$0.641$ 	&$0.634$ 	&$0.625$\\
\emph{ODR}	&$0.760$ 	&$0.815$ 	&$0.830$ 	&$0.834$ 	&$0.830$ 	&$0.828$ 	&$0.827$ 	&$0.823$\\
\emph{PD}	&$0.755$	&$0.800$	&$0.789$	&$0.784$	&$0.782$	&$0.779$	&$0.770$	&$0.765$\\
\emph{Semeion}	&$0.465$ 	&$0.638$ 	&$0.656$ 	&$0.655$ 	&$0.655$ 	&$0.658$ 	&$0.649$ 	&$0.644$\\
\emph{SPF}	&$0.163$	&$0.176$	&$0.167$	&$0.160$	&$0.157$	&$0.156$	&$0.154$	&$0.153$\\
\emph{Texture}	&$0.711$	&$0.760$	&$0.762$	&$0.752$	&$0.747$	&$0.745$	&$0.732$	&$0.728$\\
\emph{VS}	&$0.158$	&$0.161$	&$0.171$	&$0.171$	&$0.169$	&$0.169$	&$0.168$	&$0.166$\\
\emph{USPS}	&$0.526$ 	&$0.598$ 	&$0.649$ 	&$0.642$ 	&$0.639$ 	&$0.634$ 	&$0.618$ 	&$0.607$\\
\bottomrule
\end{tabular}
\end{table}

\begin{figure}[!tb]
\begin{center}
{
{\includegraphics[width=0.99\linewidth]{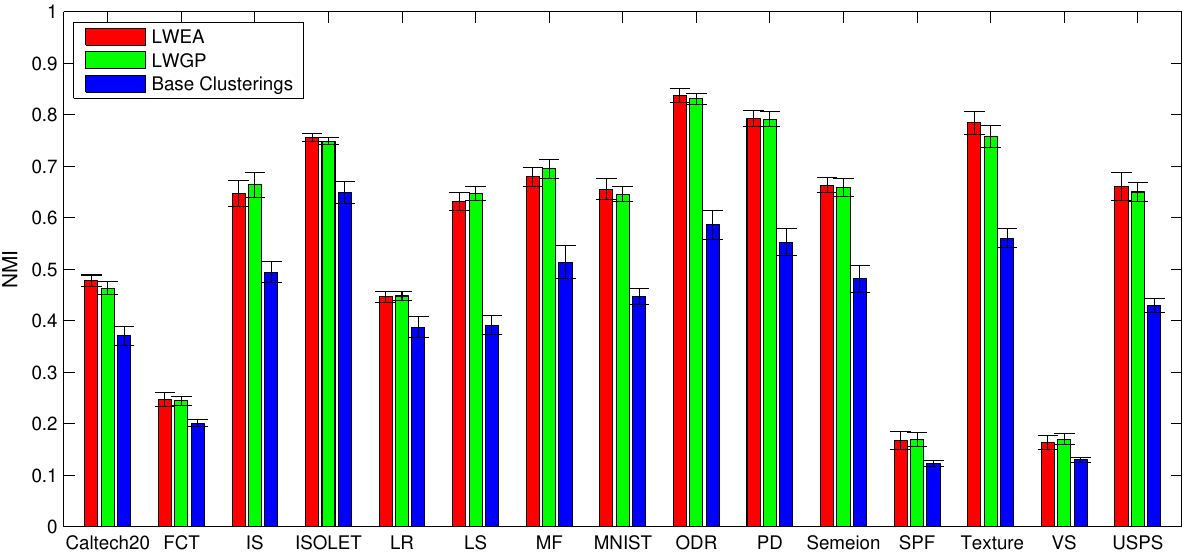}}} 
\caption{Average performances in terms of NMI of our methods and the base clusterings over $100$ runs.}
\label{fig:base_comp}
\end{center}\vskip -0.12in
\end{figure}

To evaluate the consensus performances of different algorithms over various ensembles, we construct a pool of a large number of candidate base clusterings. Each of the candidate clusterings is produced by the $k$-means algorithm with the number of clusters $k$ randomly selected in the interval of $[2, \sqrt{N}]$, where $N$ is the number of objects in the dataset. In this work, a pool of 100 candidate clusterings are randomly generated for each benchmark dataset.

\begin{table*}[!t]\footnotesize
\centering \vskip -0.1 in
\caption{Average performances (w.r.t. NMI) over 100 runs by different ensemble clustering methods
(The best two scores in each column are highlighted in bold).}\vskip -0.05 in
\label{table:compare_ce_nmi}
\begin{tabular}{m{1.285cm}<{\centering}|m{1.18cm}<{\centering}m{1.29cm}<{\centering}|m{1.18cm}<{\centering}m{1.29cm}<{\centering}|m{1.18cm}<{\centering}m{1.29cm}<{\centering}|m{1.18cm}<{\centering}m{1.29cm}<{\centering}|m{1.18cm}<{\centering}m{1.29cm}<{\centering}}
\toprule
\multirow{2}{*}{\emph{Method}}&\multicolumn{2}{c|}{\emph{Caltech20}} &\multicolumn{2}{c|}{\emph{FCT}} &\multicolumn{2}{c|}{\emph{IS}} &\multicolumn{2}{c|}{\emph{ISOLET}} &\multicolumn{2}{c}{\emph{LR}} \\
\cline{2-11}
&Best-$k$	&True-$k$&Best-$k$	&True-$k$&Best-$k$	&True-$k$&Best-$k$	&True-$k$&Best-$k$	&True-$k$\\
\hline
LWEA	&\textbf{0.478}$_{\pm0.011}$	&\textbf{0.452}$_{\pm0.011}$	&\textbf{0.247}$_{\pm0.014}$	&\textbf{0.231}$_{\pm0.024}$	&\textbf{0.647}$_{\pm0.026}$	&\textbf{0.621}$_{\pm0.026}$	&\textbf{0.756}$_{\pm0.008}$	&\textbf{0.745}$_{\pm0.011}$	&\textbf{0.446}$_{\pm0.011}$	&\textbf{0.416}$_{\pm0.017}$\\
LWGP	&\textbf{0.463}$_{\pm0.013}$	&{0.430}$_{\pm0.014}$	&\textbf{0.244}$_{\pm0.009}$	&{0.200}$_{\pm0.031}$	&\textbf{0.664}$_{\pm0.024}$	&\textbf{0.629}$_{\pm0.029}$	&{0.749}$_{\pm0.007}$	&\textbf{0.743}$_{\pm0.010}$	&\textbf{0.448}$_{\pm0.008}$	&\textbf{0.411}$_{\pm0.013}$\\
\hline
SEC	&0.401$_{\pm0.014}$	&0.377$_{\pm0.015}$	&0.218$_{\pm0.011}$	&0.148$_{\pm0.038}$	&0.591$_{\pm0.027}$	&0.437$_{\pm0.092}$	&0.699$_{\pm0.014}$	&0.651$_{\pm0.034}$	&0.408$_{\pm0.010}$	&0.299$_{\pm0.021}$\\
KCC	&0.405$_{\pm0.011}$	&0.379$_{\pm0.013}$	&0.216$_{\pm0.011}$	&0.157$_{\pm0.034}$	&0.594$_{\pm0.028}$	&0.506$_{\pm0.066}$	&0.695$_{\pm0.011}$	&0.669$_{\pm0.019}$	&0.407$_{\pm0.006}$	&0.327$_{\pm0.013}$\\
TOME	&0.399$_{\pm0.015}$	&0.382$_{\pm0.013}$	&0.228$_{\pm0.016}$	&0.199$_{\pm0.032}$	&0.574$_{\pm0.034}$	&0.476$_{\pm0.055}$	&0.712$_{\pm0.013}$	&0.691$_{\pm0.015}$	&0.427$_{\pm0.012}$	&0.353$_{\pm0.017}$\\
GP-MGLA	&0.454$_{\pm0.015}$	&0.415$_{\pm0.011}$	&0.237$_{\pm0.009}$	&0.190$_{\pm0.018}$	&0.636$_{\pm0.022}$	&0.619$_{\pm0.019}$	&0.747$_{\pm0.006}$	&0.740$_{\pm0.008}$	&0.440$_{\pm0.005}$	&0.392$_{\pm0.011}$\\
WEAC	&0.461$_{\pm0.014}$	&0.435$_{\pm0.012}$	&0.232$_{\pm0.014}$	&0.206$_{\pm0.021}$	&0.619$_{\pm0.020}$	&0.600$_{\pm0.021}$	&0.749$_{\pm0.008}$	&0.734$_{\pm0.018}$	&0.435$_{\pm0.008}$	&0.384$_{\pm0.018}$\\
WCT	&0.462$_{\pm0.012}$	&\textbf{0.447}$_{\pm0.014}$	&0.237$_{\pm0.013}$	&0.211$_{\pm0.021}$	&0.630$_{\pm0.019}$	&0.603$_{\pm0.019}$	&\textbf{0.755}$_{\pm0.008}$	&0.719$_{\pm0.029}$	&0.434$_{\pm0.010}$	&0.384$_{\pm0.021}$\\
EAC	&0.456$_{\pm0.015}$	&0.434$_{\pm0.013}$	&0.229$_{\pm0.014}$	&0.203$_{\pm0.022}$	&0.620$_{\pm0.021}$	&0.599$_{\pm0.022}$	&0.749$_{\pm0.008}$	&0.730$_{\pm0.021}$	&0.431$_{\pm0.009}$	&0.365$_{\pm0.021}$\\
HBGF	&0.453$_{\pm0.013}$	&0.416$_{\pm0.010}$	&0.233$_{\pm0.008}$	&0.188$_{\pm0.021}$	&0.627$_{\pm0.023}$	&0.609$_{\pm0.025}$	&0.748$_{\pm0.006}$	&0.742$_{\pm0.008}$	&0.440$_{\pm0.004}$	&0.385$_{\pm0.013}$\\
MCLA	&0.413$_{\pm0.012}$	&0.339$_{\pm0.070}$	&0.232$_{\pm0.013}$	&\textbf{0.218}$_{\pm0.025}$	&0.632$_{\pm0.030}$	&\textbf{0.621}$_{\pm0.035}$	&0.719$_{\pm0.021}$	&0.665$_{\pm0.015}$	&0.404$_{\pm0.018}$	&0.351$_{\pm0.017}$\\
HGPA	&0.363$_{\pm0.021}$	&0.316$_{\pm0.028}$	&0.172$_{\pm0.011}$	&0.118$_{\pm0.035}$	&0.501$_{\pm0.033}$	&0.446$_{\pm0.071}$	&0.637$_{\pm0.023}$	&0.486$_{\pm0.042}$	&0.359$_{\pm0.009}$	&0.174$_{\pm0.019}$\\
CSPA	&0.381$_{\pm0.013}$	&0.349$_{\pm0.010}$	&0.220$_{\pm0.017}$	&0.207$_{\pm0.019}$	&0.611$_{\pm0.025}$	&0.610$_{\pm0.028}$	&0.670$_{\pm0.014}$	&0.629$_{\pm0.014}$	&0.347$_{\pm0.080}$	&0.280$_{\pm0.065}$\\
\bottomrule
\toprule
\multirow{2}{*}{\emph{Method}}&\multicolumn{2}{c|}{\emph{LS}} &\multicolumn{2}{c|}{\emph{MF}} &\multicolumn{2}{c|}{\emph{MNIST}} &\multicolumn{2}{c|}{\emph{ODR}} &\multicolumn{2}{c}{\emph{PD}} \\
\cline{2-11}
&Best-$k$	&True-$k$&Best-$k$	&True-$k$&Best-$k$	&True-$k$&Best-$k$	&True-$k$&Best-$k$	&True-$k$\\
\hline
LWEA&\textbf{0.632}$_{\pm0.018}$	&{0.616}$_{\pm0.027}$	&{0.679}$_{\pm0.019}$	&{0.659}$_{\pm0.021}$	&\textbf{0.655}$_{\pm0.020}$	&\textbf{0.646}$_{\pm0.022}$	&\textbf{0.838}$_{\pm0.014}$	&\textbf{0.829}$_{\pm0.018}$	&\textbf{0.793}$_{\pm0.015}$	&{0.769}$_{\pm0.022}$\\
LWGP	&\textbf{0.648}$_{\pm0.014}$	&\textbf{0.644}$_{\pm0.019}$	&\textbf{0.695}$_{\pm0.019}$	&\textbf{0.682}$_{\pm0.026}$	&\textbf{0.646}$_{\pm0.015}$	&\textbf{0.635}$_{\pm0.017}$	&\textbf{0.831}$_{\pm0.011}$	&\textbf{0.816}$_{\pm0.015}$	&{0.792}$_{\pm0.015}$	&\textbf{0.774}$_{\pm0.021}$\\
\hline
SEC	&0.478$_{\pm0.034}$	&0.380$_{\pm0.074}$	&0.478$_{\pm0.034}$	&0.380$_{\pm0.074}$	&0.506$_{\pm0.022}$	&0.423$_{\pm0.049}$	&0.697$_{\pm0.027}$	&0.604$_{\pm0.065}$	&0.653$_{\pm0.025}$	&0.552$_{\pm0.065}$\\
KCC	&0.494$_{\pm0.033}$	&0.442$_{\pm0.061}$	&0.494$_{\pm0.033}$	&0.442$_{\pm0.061}$	&0.523$_{\pm0.018}$	&0.480$_{\pm0.037}$	&0.719$_{\pm0.022}$	&0.667$_{\pm0.039}$	&0.664$_{\pm0.023}$	&0.598$_{\pm0.053}$\\
TOME	&0.521$_{\pm0.031}$	&0.510$_{\pm0.038}$	&\textbf{0.701}$_{\pm0.025}$	&\textbf{0.687}$_{\pm0.029}$	&0.584$_{\pm0.025}$	&0.553$_{\pm0.034}$	&0.814$_{\pm0.020}$	&0.794$_{\pm0.030}$	&\textbf{0.801}$_{\pm0.020}$	&\textbf{0.789}$_{\pm0.027}$\\
GP-MGLA	&0.629$_{\pm0.014}$	&\textbf{0.619}$_{\pm0.020}$	&0.661$_{\pm0.028}$	&0.638$_{\pm0.031}$	&0.628$_{\pm0.027}$	&0.616$_{\pm0.026}$	&0.825$_{\pm0.018}$	&0.813$_{\pm0.020}$	&0.767$_{\pm0.021}$	&0.735$_{\pm0.031}$\\
WEAC	&0.613$_{\pm0.031}$	&0.601$_{\pm0.048}$	&0.638$_{\pm0.029}$	&0.609$_{\pm0.038}$	&0.623$_{\pm0.026}$	&0.615$_{\pm0.027}$	&0.820$_{\pm0.019}$	&0.801$_{\pm0.019}$	&0.757$_{\pm0.021}$	&0.716$_{\pm0.030}$\\
WCT	&0.622$_{\pm0.026}$	&0.602$_{\pm0.055}$	&0.650$_{\pm0.028}$	&0.614$_{\pm0.043}$	&0.634$_{\pm0.024}$	&0.613$_{\pm0.034}$	&0.822$_{\pm0.017}$	&0.798$_{\pm0.027}$	&0.766$_{\pm0.017}$	&0.706$_{\pm0.038}$\\
EAC	&0.597$_{\pm0.044}$	&0.559$_{\pm0.086}$	&0.632$_{\pm0.028}$	&0.597$_{\pm0.038}$	&0.611$_{\pm0.026}$	&0.592$_{\pm0.037}$	&0.807$_{\pm0.023}$	&0.781$_{\pm0.035}$	&0.751$_{\pm0.020}$	&0.697$_{\pm0.036}$\\
HBGF	&0.630$_{\pm0.021}$	&0.618$_{\pm0.035}$	&0.658$_{\pm0.026}$	&0.636$_{\pm0.031}$	&0.618$_{\pm0.029}$	&0.607$_{\pm0.030}$	&0.819$_{\pm0.019}$	&0.810$_{\pm0.018}$	&0.760$_{\pm0.019}$	&0.730$_{\pm0.024}$\\
MCLA	&0.547$_{\pm0.025}$	&0.518$_{\pm0.036}$	&0.653$_{\pm0.033}$	&0.627$_{\pm0.065}$	&0.574$_{\pm0.030}$	&0.554$_{\pm0.040}$	&0.792$_{\pm0.030}$	&0.775$_{\pm0.038}$	&0.694$_{\pm0.026}$	&0.678$_{\pm0.035}$\\
HGPA	&0.386$_{\pm0.031}$	&0.312$_{\pm0.066}$	&0.538$_{\pm0.040}$	&0.479$_{\pm0.078}$	&0.426$_{\pm0.031}$	&0.296$_{\pm0.077}$	&0.621$_{\pm0.042}$	&0.409$_{\pm0.089}$	&0.560$_{\pm0.040}$	&0.308$_{\pm0.060}$\\
CSPA	&0.522$_{\pm0.037}$	&0.485$_{\pm0.040}$	&0.625$_{\pm0.027}$	&0.617$_{\pm0.030}$	&0.527$_{\pm0.040}$	&0.521$_{\pm0.042}$	&0.741$_{\pm0.049}$	&0.738$_{\pm0.052}$	&0.661$_{\pm0.032}$	&0.659$_{\pm0.035}$\\
\bottomrule
\toprule
\multirow{2}{*}{\emph{Method}}&\multicolumn{2}{c|}{\emph{Semeion}} &\multicolumn{2}{c|}{\emph{SPF}} &\multicolumn{2}{c|}{\emph{Texture}} &\multicolumn{2}{c|}{\emph{VS}} &\multicolumn{2}{c}{\emph{USPS}} \\
\cline{2-11}
&Best-$k$	&True-$k$&Best-$k$	&True-$k$&Best-$k$	&True-$k$&Best-$k$	&True-$k$&Best-$k$	&True-$k$\\
\hline
LWEA&\textbf{0.663}$_{\pm0.015}$	&\textbf{0.655}$_{\pm0.017}$	&\textbf{0.167}$_{\pm0.017}$	&{0.151}$_{\pm0.029}$	&\textbf{0.784}$_{\pm0.023}$	&\textbf{0.778}$_{\pm0.028}$	&{0.163}$_{\pm0.014}$	&\textbf{0.133}$_{\pm0.010}$	&\textbf{0.661}$_{\pm0.027}$	&\textbf{0.633}$_{\pm0.032}$\\
LWGP	&\textbf{0.658}$_{\pm0.017}$	&\textbf{0.642}$_{\pm0.024}$	&\textbf{0.169}$_{\pm0.014}$	&\textbf{0.152}$_{\pm0.023}$	&\textbf{0.757}$_{\pm0.021}$	&\textbf{0.743}$_{\pm0.024}$	&\textbf{0.170}$_{\pm0.011}$	&\textbf{0.132}$_{\pm0.012}$	&\textbf{0.650}$_{\pm0.019}$	&\textbf{0.614}$_{\pm0.020}$\\
\hline
SEC	&0.544$_{\pm0.025}$	&0.466$_{\pm0.046}$	&0.132$_{\pm0.009}$	&0.073$_{\pm0.025}$	&0.642$_{\pm0.020}$	&0.533$_{\pm0.053}$	&0.148$_{\pm0.011}$	&0.116$_{\pm0.029}$	&0.477$_{\pm0.021}$	&0.372$_{\pm0.049}$\\
KCC	&0.551$_{\pm0.019}$	&0.507$_{\pm0.033}$	&0.130$_{\pm0.008}$	&0.079$_{\pm0.028}$	&0.648$_{\pm0.018}$	&0.569$_{\pm0.042}$	&0.146$_{\pm0.012}$	&0.126$_{\pm0.025}$	&0.503$_{\pm0.015}$	&0.450$_{\pm0.041}$\\
TOME	&0.603$_{\pm0.026}$	&0.575$_{\pm0.034}$	&0.166$_{\pm0.011}$	&\textbf{0.153}$_{\pm0.015}$	&0.740$_{\pm0.026}$	&0.646$_{\pm0.051}$	&0.144$_{\pm0.012}$	&0.104$_{\pm0.038}$	&0.601$_{\pm0.028}$	&0.573$_{\pm0.035}$\\
GP-MGLA	&0.640$_{\pm0.022}$	&0.623$_{\pm0.026}$	&0.156$_{\pm0.009}$	&0.137$_{\pm0.016}$	&0.725$_{\pm0.024}$	&0.717$_{\pm0.025}$	&0.163$_{\pm0.011}$	&0.127$_{\pm0.011}$	&0.609$_{\pm0.030}$	&0.597$_{\pm0.033}$\\
WEAC	&0.642$_{\pm0.021}$	&0.628$_{\pm0.026}$	&0.154$_{\pm0.013}$	&0.124$_{\pm0.024}$	&0.730$_{\pm0.027}$	&0.713$_{\pm0.030}$	&0.163$_{\pm0.012}$	&0.130$_{\pm0.015}$	&0.602$_{\pm0.036}$	&0.583$_{\pm0.038}$\\
WCT	&0.653$_{\pm0.020}$	&0.628$_{\pm0.034}$	&0.161$_{\pm0.012}$	&0.131$_{\pm0.022}$	&0.742$_{\pm0.027}$	&0.722$_{\pm0.032}$	&\textbf{0.164}$_{\pm0.013}$	&0.126$_{\pm0.012}$	&0.604$_{\pm0.029}$	&0.579$_{\pm0.036}$\\
EAC	&0.637$_{\pm0.023}$	&0.616$_{\pm0.033}$	&0.153$_{\pm0.013}$	&0.114$_{\pm0.024}$	&0.717$_{\pm0.027}$	&0.695$_{\pm0.032}$	&0.160$_{\pm0.012}$	&0.129$_{\pm0.014}$	&0.582$_{\pm0.034}$	&0.555$_{\pm0.047}$\\
HBGF	&0.637$_{\pm0.023}$	&0.621$_{\pm0.026}$	&0.157$_{\pm0.008}$	&0.129$_{\pm0.018}$	&0.719$_{\pm0.023}$	&0.706$_{\pm0.027}$	&0.162$_{\pm0.011}$	&0.127$_{\pm0.012}$	&0.597$_{\pm0.030}$	&0.581$_{\pm0.031}$\\
MCLA	&0.595$_{\pm0.022}$	&0.572$_{\pm0.038}$	&0.139$_{\pm0.012}$	&0.095$_{\pm0.029}$	&0.701$_{\pm0.019}$	&0.687$_{\pm0.027}$	&0.150$_{\pm0.012}$	&0.130$_{\pm0.026}$	&0.553$_{\pm0.023}$	&0.531$_{\pm0.036}$\\
HGPA	&0.503$_{\pm0.026}$	&0.452$_{\pm0.053}$	&0.121$_{\pm0.012}$	&0.093$_{\pm0.028}$	&0.495$_{\pm0.029}$	&0.346$_{\pm0.061}$	&0.127$_{\pm0.016}$	&0.096$_{\pm0.032}$	&0.373$_{\pm0.027}$	&0.120$_{\pm0.046}$\\
CSPA	&0.549$_{\pm0.041}$	&0.537$_{\pm0.050}$	&0.115$_{\pm0.007}$	&0.072$_{\pm0.014}$	&0.658$_{\pm0.022}$	&0.655$_{\pm0.022}$	&0.137$_{\pm0.017}$	&0.131$_{\pm0.022}$	&0.513$_{\pm0.046}$	&0.501$_{\pm0.054}$\\
\bottomrule
\end{tabular}
\end{table*}

\begin{table*}[!t]\footnotesize
\centering \vskip -0.05 in
\caption{The number of times that our method is \textbf{(significantly better than / comparable to / significantly worse than)} a baseline method by statistical test (t-test with $p<0.05$) on the results in Table~\ref{table:compare_ce_nmi}.}\vskip -0.05 in
\label{table:compare_ce2_nmi}
\begin{tabular}{m{1.01cm}<{\centering}m{1.01cm}<{\centering}m{1.01cm}<{\centering}m{1.01cm}<{\centering}m{1.30cm}<{\centering}m{1.01cm}<{\centering}m{1.01cm}<{\centering}m{1.01cm}<{\centering}m{1.01cm}<{\centering}m{1.01cm}<{\centering}m{1.01cm}<{\centering}m{1.01cm}<{\centering}}
\toprule
        &SEC   &KCC    &TOME   &GP-MGLA    &WEAC   &WCT    &EAC    &HBGF   &MCLA   &HGPA   &CSPA\\
\midrule
LWEA	&(\textbf{30}$/$0$/$0)	&(\textbf{30}$/$0$/$0)	&(\textbf{24}$/$2$/$4)	&(\textbf{26}$/$4$/$0)	&(\textbf{28}$/$2$/$0)	&(\textbf{28}$/$2$/$0)	&(\textbf{29}$/$1$/$0)	&(\textbf{27}$/$3$/$0)	&(\textbf{28}$/$2$/$0)	&(\textbf{30}$/$0$/$0)	&(\textbf{29}$/$1$/$0)\\
\midrule
LWGP	&(\textbf{30}$/$0$/$0)	&(\textbf{30}$/$0$/$0)	&(\textbf{23}$/$4$/$3)	&(\textbf{29}$/$1$/$0)	&(\textbf{25}$/$4$/$1)	&(\textbf{26}$/$1$/$3)	&(\textbf{26}$/$3$/$1)	&(\textbf{28}$/$2$/$0)	&(\textbf{27}$/$2$/$1)	&(\textbf{30}$/$0$/$0)	&(\textbf{28}$/$2$/$0)\\
\bottomrule
\end{tabular}
\vskip -0.05in
\end{table*}

With the base clustering pool generated, to rule out the factor of \emph{getting lucky occasionally} and provide a fair comparison, the proposed methods and the baseline methods are evaluated by their average performances over a large number of runs, where the clustering ensemble for each run is constructed by randomly choosing $M$ base clusterings from the pool. Typically, the ensemble size $M=10$ is used. The consensus performances of different methods with varying ensemble sizes are also evaluated in the following of this paper (see Section~\ref{sec:ensize}).

\subsection{Choices of Parameter $\theta$}
\label{sec:para_anal}

\begin{table*}[!t]\footnotesize
\centering \vskip -0.1 in
\caption{Average performances (w.r.t. ARI) over 100 runs by different ensemble clustering methods
(The best two scores in each column are highlighted in bold).}\vskip -0.05 in
\label{table:compare_ce_ari}
\begin{tabular}{m{1.285cm}<{\centering}|m{1.18cm}<{\centering}m{1.29cm}<{\centering}|m{1.18cm}<{\centering}m{1.29cm}<{\centering}|m{1.18cm}<{\centering}m{1.29cm}<{\centering}|m{1.18cm}<{\centering}m{1.29cm}<{\centering}|m{1.18cm}<{\centering}m{1.29cm}<{\centering}}
\toprule
\multirow{2}{*}{\emph{Method}}&\multicolumn{2}{c|}{\emph{Caltech20}} &\multicolumn{2}{c|}{\emph{FCT}} &\multicolumn{2}{c|}{\emph{IS}} &\multicolumn{2}{c|}{\emph{ISOLET}} &\multicolumn{2}{c}{\emph{LR}} \\
\cline{2-11}
&Best-$k$	&True-$k$&Best-$k$	&True-$k$&Best-$k$	&True-$k$&Best-$k$	&True-$k$&Best-$k$	&True-$k$\\
\hline
LWEA	&\textbf{0.448}$_{\pm0.037}$	&\textbf{0.352}$_{\pm0.036}$	&\textbf{0.161}$_{\pm0.022}$	&\textbf{0.129}$_{\pm0.019}$	&\textbf{0.586}$_{\pm0.027}$	&\textbf{0.522}$_{\pm0.031}$	&\textbf{0.572}$_{\pm0.017}$	&\textbf{0.555}$_{\pm0.021}$	&\textbf{0.223}$_{\pm0.013}$	&\textbf{0.200}$_{\pm0.016}$\\
LWGP	&\textbf{0.399}$_{\pm0.035}$	&{0.267}$_{\pm0.032}$	&\textbf{0.152}$_{\pm0.012}$	&{0.117}$_{\pm0.026}$	&\textbf{0.573}$_{\pm0.031}$	&\textbf{0.529}$_{\pm0.039}$	&{0.536}$_{\pm0.017}$	&\textbf{0.518}$_{\pm0.024}$	&\textbf{0.188}$_{\pm0.007}$	&\textbf{0.162}$_{\pm0.013}$\\
\hline
SEC	&0.331$_{\pm0.056}$	&0.221$_{\pm0.043}$	&0.119$_{\pm0.019}$	&0.078$_{\pm0.038}$	&0.497$_{\pm0.041}$	&0.300$_{\pm0.109}$	&0.469$_{\pm0.024}$	&0.390$_{\pm0.066}$	&0.157$_{\pm0.010}$	&0.097$_{\pm0.020}$\\
KCC	&0.343$_{\pm0.047}$	&0.225$_{\pm0.037}$	&0.123$_{\pm0.017}$	&0.084$_{\pm0.032}$	&0.508$_{\pm0.041}$	&0.395$_{\pm0.079}$	&0.467$_{\pm0.022}$	&0.420$_{\pm0.038}$	&0.160$_{\pm0.007}$	&0.122$_{\pm0.012}$\\
TOME	&0.270$_{\pm0.052}$	&0.169$_{\pm0.022}$	&0.127$_{\pm0.023}$	&0.110$_{\pm0.030}$	&0.386$_{\pm0.046}$	&0.266$_{\pm0.081}$	&0.439$_{\pm0.029}$	&0.417$_{\pm0.032}$	&0.138$_{\pm0.011}$	&0.116$_{\pm0.014}$\\
GP-MGLA	&0.376$_{\pm0.043}$	&0.238$_{\pm0.028}$	&0.151$_{\pm0.009}$	&0.099$_{\pm0.016}$	&0.552$_{\pm0.025}$	&0.521$_{\pm0.029}$	&0.531$_{\pm0.017}$	&0.507$_{\pm0.021}$	&0.175$_{\pm0.005}$	&0.140$_{\pm0.012}$\\
WEAC	&0.395$_{\pm0.036}$	&0.302$_{\pm0.032}$	&0.147$_{\pm0.020}$	&0.120$_{\pm0.020}$	&0.552$_{\pm0.027}$	&0.497$_{\pm0.031}$	&\textbf{0.539}$_{\pm0.018}$	&\textbf{0.518}$_{\pm0.028}$	&0.177$_{\pm0.008}$	&0.148$_{\pm0.014}$\\
WCT	&0.392$_{\pm0.023}$	&\textbf{0.334}$_{\pm0.036}$	&0.151$_{\pm0.020}$	&\textbf{0.127}$_{\pm0.024}$	&0.560$_{\pm0.025}$	&0.505$_{\pm0.031}$	&0.546$_{\pm0.018}$	&0.511$_{\pm0.037}$	&0.175$_{\pm0.009}$	&0.138$_{\pm0.012}$\\
EAC	&0.390$_{\pm0.037}$	&0.308$_{\pm0.032}$	&0.144$_{\pm0.020}$	&0.122$_{\pm0.020}$	&0.550$_{\pm0.029}$	&0.491$_{\pm0.037}$	&0.536$_{\pm0.019}$	&0.516$_{\pm0.030}$	&0.171$_{\pm0.007}$	&0.134$_{\pm0.017}$\\
HBGF	&0.360$_{\pm0.025}$	&0.235$_{\pm0.023}$	&0.147$_{\pm0.011}$	&0.097$_{\pm0.018}$	&0.550$_{\pm0.029}$	&0.509$_{\pm0.030}$	&0.529$_{\pm0.020}$	&0.511$_{\pm0.020}$	&0.170$_{\pm0.004}$	&0.138$_{\pm0.010}$\\
MCLA	&0.347$_{\pm0.031}$	&0.161$_{\pm0.077}$	&0.145$_{\pm0.012}$	&0.119$_{\pm0.015}$	&0.517$_{\pm0.042}$	&0.480$_{\pm0.043}$	&0.518$_{\pm0.026}$	&0.478$_{\pm0.107}$	&0.186$_{\pm0.013}$	&0.161$_{\pm0.027}$\\
HGPA	&0.242$_{\pm0.036}$	&0.148$_{\pm0.018}$	&0.081$_{\pm0.010}$	&0.062$_{\pm0.017}$	&0.362$_{\pm0.034}$	&0.315$_{\pm0.061}$	&0.379$_{\pm0.033}$	&0.365$_{\pm0.037}$	&0.130$_{\pm0.007}$	&0.116$_{\pm0.010}$\\
CSPA	&0.319$_{\pm0.029}$	&0.170$_{\pm0.006}$	&0.151$_{\pm0.008}$	&0.118$_{\pm0.009}$	&0.468$_{\pm0.036}$	&0.458$_{\pm0.042}$	&0.486$_{\pm0.037}$	&0.482$_{\pm0.042}$	&0.107$_{\pm0.064}$	&0.097$_{\pm0.063}$\\
\bottomrule
\toprule
\multirow{2}{*}{\emph{Method}}&\multicolumn{2}{c|}{\emph{LS}} &\multicolumn{2}{c|}{\emph{MF}} &\multicolumn{2}{c|}{\emph{MNIST}} &\multicolumn{2}{c|}{\emph{ODR}} &\multicolumn{2}{c}{\emph{PD}} \\
\cline{2-11}
&Best-$k$	&True-$k$&Best-$k$	&True-$k$&Best-$k$	&True-$k$&Best-$k$	&True-$k$&Best-$k$	&True-$k$\\
\hline
LWEA	&\textbf{0.614}$_{\pm0.037}$	&\textbf{0.568}$_{\pm0.054}$	&\textbf{0.572}$_{\pm0.026}$	&{0.525}$_{\pm0.030}$	&\textbf{0.572}$_{\pm0.032}$	&\textbf{0.550}$_{\pm0.037}$	&\textbf{0.836}$_{\pm0.017}$	&\textbf{0.782}$_{\pm0.032}$	&\textbf{0.747}$_{\pm0.017}$	&\textbf{0.675}$_{\pm0.029}$\\
LWGP	&{0.598}$_{\pm0.013}$	&\textbf{0.580}$_{\pm0.032}$	&\textbf{0.591}$_{\pm0.021}$	&\textbf{0.562}$_{\pm0.035}$	&{0.540}$_{\pm0.022}$	&\textbf{0.512}$_{\pm0.026}$	&\textbf{0.823}$_{\pm0.019}$	&\textbf{0.763}$_{\pm0.026}$	&\textbf{0.739}$_{\pm0.019}$	&\textbf{0.675}$_{\pm0.040}$\\
\hline
SEC	&0.370$_{\pm0.054}$	&0.235$_{\pm0.093}$	&0.465$_{\pm0.027}$	&0.361$_{\pm0.072}$	&0.369$_{\pm0.037}$	&0.263$_{\pm0.070}$	&0.602$_{\pm0.047}$	&0.427$_{\pm0.100}$	&0.532$_{\pm0.060}$	&0.373$_{\pm0.093}$\\
KCC	&0.399$_{\pm0.051}$	&0.304$_{\pm0.078}$	&0.474$_{\pm0.025}$	&0.402$_{\pm0.042}$	&0.400$_{\pm0.031}$	&0.333$_{\pm0.056}$	&0.642$_{\pm0.042}$	&0.525$_{\pm0.069}$	&0.551$_{\pm0.046}$	&0.438$_{\pm0.082}$\\
TOME	&0.423$_{\pm0.056}$	&0.362$_{\pm0.053}$	&0.571$_{\pm0.035}$	&\textbf{0.549}$_{\pm0.046}$	&0.403$_{\pm0.037}$	&0.385$_{\pm0.045}$	&0.738$_{\pm0.034}$	&0.701$_{\pm0.055}$	&0.737$_{\pm0.031}$	&\textbf{0.686}$_{\pm0.047}$\\
GP-MGLA	&0.600$_{\pm0.033}$	&0.538$_{\pm0.038}$	&0.558$_{\pm0.023}$	&0.513$_{\pm0.032}$	&\textbf{0.558}$_{\pm0.023}$	&0.511$_{\pm0.032}$	&0.808$_{\pm0.027}$	&0.760$_{\pm0.033}$	&0.700$_{\pm0.025}$	&0.628$_{\pm0.049}$\\
WEAC	&0.590$_{\pm0.063}$	&0.538$_{\pm0.084}$	&0.531$_{\pm0.023}$	&0.467$_{\pm0.037}$	&0.531$_{\pm0.023}$	&0.467$_{\pm0.037}$	&0.797$_{\pm0.036}$	&0.731$_{\pm0.031}$	&0.695$_{\pm0.034}$	&0.593$_{\pm0.045}$\\
WCT	&\textbf{0.606}$_{\pm0.050}$	&0.549$_{\pm0.077}$	&0.539$_{\pm0.024}$	&0.475$_{\pm0.034}$	&0.539$_{\pm0.024}$	&0.475$_{\pm0.034}$	&0.816$_{\pm0.029}$	&0.729$_{\pm0.041}$	&0.721$_{\pm0.018}$	&0.579$_{\pm0.052}$\\
EAC	&0.571$_{\pm0.073}$	&0.486$_{\pm0.115}$	&0.526$_{\pm0.022}$	&0.455$_{\pm0.037}$	&0.526$_{\pm0.022}$	&0.455$_{\pm0.037}$	&0.779$_{\pm0.041}$	&0.698$_{\pm0.058}$	&0.686$_{\pm0.031}$	&0.566$_{\pm0.052}$\\
HBGF	&0.586$_{\pm0.046}$	&0.540$_{\pm0.062}$	&0.554$_{\pm0.028}$	&0.505$_{\pm0.036}$	&0.498$_{\pm0.032}$	&0.479$_{\pm0.038}$	&0.795$_{\pm0.027}$	&0.751$_{\pm0.033}$	&0.690$_{\pm0.026}$	&0.621$_{\pm0.039}$\\
MCLA	&0.496$_{\pm0.051}$	&0.443$_{\pm0.053}$	&0.543$_{\pm0.045}$	&0.508$_{\pm0.084}$	&0.451$_{\pm0.042}$	&0.428$_{\pm0.055}$	&0.727$_{\pm0.054}$	&0.706$_{\pm0.062}$	&0.594$_{\pm0.035}$	&0.551$_{\pm0.057}$\\
HGPA	&0.275$_{\pm0.040}$	&0.228$_{\pm0.055}$	&0.378$_{\pm0.043}$	&0.315$_{\pm0.074}$	&0.260$_{\pm0.032}$	&0.185$_{\pm0.052}$	&0.476$_{\pm0.066}$	&0.281$_{\pm0.068}$	&0.423$_{\pm0.055}$	&0.197$_{\pm0.050}$\\
CSPA	&0.460$_{\pm0.062}$	&0.402$_{\pm0.048}$	&0.513$_{\pm0.039}$	&0.507$_{\pm0.045}$	&0.420$_{\pm0.048}$	&0.411$_{\pm0.053}$	&0.676$_{\pm0.076}$	&0.675$_{\pm0.076}$	&0.559$_{\pm0.042}$	&0.551$_{\pm0.051}$\\
\bottomrule
\toprule
\multirow{2}{*}{\emph{Method}}&\multicolumn{2}{c|}{\emph{Semeion}} &\multicolumn{2}{c|}{\emph{SPF}} &\multicolumn{2}{c|}{\emph{Texture}} &\multicolumn{2}{c|}{\emph{VS}} &\multicolumn{2}{c}{\emph{USPS}} \\
\cline{2-11}
&Best-$k$	&True-$k$&Best-$k$	&True-$k$&Best-$k$	&True-$k$&Best-$k$	&True-$k$&Best-$k$	&True-$k$\\
\hline
LWEA	&\textbf{0.548}$_{\pm0.022}$	&\textbf{0.539}$_{\pm0.024}$	&{0.097}$_{\pm0.020}$	&\textbf{0.084}$_{\pm0.026}$	&\textbf{0.712}$_{\pm0.028}$	&\textbf{0.689}$_{\pm0.046}$	&\textbf{0.126}$_{\pm0.012}$	&\textbf{0.116}$_{\pm0.014}$	&\textbf{0.559}$_{\pm0.046}$	&\textbf{0.512}$_{\pm0.049}$\\
LWGP	&\textbf{0.541}$_{\pm0.022}$	&\textbf{0.520}$_{\pm0.035}$	&\textbf{0.098}$_{\pm0.013}$	&{0.083}$_{\pm0.020}$	&\textbf{0.656}$_{\pm0.032}$	&\textbf{0.620}$_{\pm0.043}$	&\textbf{0.121}$_{\pm0.011}$	&{0.097}$_{\pm0.021}$	&\textbf{0.534}$_{\pm0.028}$	&\textbf{0.461}$_{\pm0.028}$\\
\hline
SEC	&0.406$_{\pm0.034}$	&0.297$_{\pm0.063}$	&0.070$_{\pm0.014}$	&0.033$_{\pm0.020}$	&0.510$_{\pm0.034}$	&0.343$_{\pm0.088}$	&0.112$_{\pm0.017}$	&0.088$_{\pm0.032}$	&0.308$_{\pm0.035}$	&0.191$_{\pm0.062}$\\
KCC	&0.423$_{\pm0.030}$	&0.355$_{\pm0.044}$	&0.066$_{\pm0.009}$	&0.041$_{\pm0.019}$	&0.526$_{\pm0.033}$	&0.401$_{\pm0.072}$	&0.113$_{\pm0.017}$	&0.100$_{\pm0.025}$	&0.348$_{\pm0.021}$	&0.278$_{\pm0.051}$\\
TOME	&0.436$_{\pm0.042}$	&0.407$_{\pm0.053}$	&\textbf{0.099}$_{\pm0.018}$	&\textbf{0.087}$_{\pm0.021}$	&0.574$_{\pm0.052}$	&0.438$_{\pm0.083}$	&0.099$_{\pm0.020}$	&0.077$_{\pm0.034}$	&0.443$_{\pm0.043}$	&0.421$_{\pm0.046}$\\
GP-MGLA	&0.513$_{\pm0.027}$	&0.488$_{\pm0.038}$	&0.086$_{\pm0.010}$	&0.069$_{\pm0.013}$	&0.609$_{\pm0.035}$	&0.585$_{\pm0.040}$	&0.118$_{\pm0.012}$	&0.097$_{\pm0.021}$	&0.493$_{\pm0.043}$	&0.449$_{\pm0.044}$\\
WEAC	&0.512$_{\pm0.027}$	&0.495$_{\pm0.036}$	&0.081$_{\pm0.013}$	&0.062$_{\pm0.016}$	&0.641$_{\pm0.034}$	&0.590$_{\pm0.042}$	&0.120$_{\pm0.015}$	&\textbf{0.103}$_{\pm0.025}$	&0.472$_{\pm0.053}$	&0.433$_{\pm0.055}$\\
WCT	&0.519$_{\pm0.026}$	&0.499$_{\pm0.038}$	&0.087$_{\pm0.013}$	&0.065$_{\pm0.019}$	&0.655$_{\pm0.031}$	&0.601$_{\pm0.047}$	&0.119$_{\pm0.013}$	&0.098$_{\pm0.023}$	&0.470$_{\pm0.048}$	&0.434$_{\pm0.046}$\\
EAC	&0.501$_{\pm0.030}$	&0.478$_{\pm0.044}$	&0.078$_{\pm0.013}$	&0.058$_{\pm0.016}$	&0.628$_{\pm0.034}$	&0.567$_{\pm0.044}$	&0.118$_{\pm0.014}$	&0.099$_{\pm0.024}$	&0.437$_{\pm0.055}$	&0.396$_{\pm0.064}$\\
HBGF	&0.511$_{\pm0.028}$	&0.491$_{\pm0.039}$	&0.083$_{\pm0.009}$	&0.064$_{\pm0.013}$	&0.601$_{\pm0.034}$	&0.572$_{\pm0.041}$	&0.118$_{\pm0.010}$	&0.099$_{\pm0.021}$	&0.474$_{\pm0.042}$	&0.425$_{\pm0.044}$\\
MCLA	&0.473$_{\pm0.036}$	&0.442$_{\pm0.058}$	&0.086$_{\pm0.015}$	&0.061$_{\pm0.024}$	&0.597$_{\pm0.028}$	&0.577$_{\pm0.039}$	&0.120$_{\pm0.018}$	&0.102$_{\pm0.023}$	&0.412$_{\pm0.031}$	&0.378$_{\pm0.047}$\\
HGPA	&0.350$_{\pm0.033}$	&0.313$_{\pm0.049}$	&0.074$_{\pm0.019}$	&0.064$_{\pm0.024}$	&0.316$_{\pm0.034}$	&0.206$_{\pm0.046}$	&0.097$_{\pm0.020}$	&0.076$_{\pm0.026}$	&0.200$_{\pm0.031}$	&0.058$_{\pm0.026}$\\
CSPA	&0.428$_{\pm0.055}$	&0.416$_{\pm0.064}$	&0.062$_{\pm0.013}$	&0.042$_{\pm0.012}$	&0.563$_{\pm0.024}$	&0.561$_{\pm0.023}$	&0.113$_{\pm0.021}$	&\textbf{0.103}$_{\pm0.021}$	&0.383$_{\pm0.058}$	&0.359$_{\pm0.065}$\\
\bottomrule
\end{tabular}
\end{table*}

\begin{table*}[!t]\footnotesize
\centering \vskip -0.08 in
\caption{The number of times that our method is \textbf{(significantly better than / comparable to / significantly worse than)} a baseline method by statistical test (t-test with $p<0.05$) on the results in Table~\ref{table:compare_ce_ari}.}\vskip -0.05 in
\label{table:compare_ce2_ari}
\begin{tabular}{m{1.01cm}<{\centering}m{1.01cm}<{\centering}m{1.01cm}<{\centering}m{1.01cm}<{\centering}m{1.30cm}<{\centering}m{1.01cm}<{\centering}m{1.01cm}<{\centering}m{1.01cm}<{\centering}m{1.01cm}<{\centering}m{1.01cm}<{\centering}m{1.01cm}<{\centering}m{1.01cm}<{\centering}}
\toprule
        &SEC   &KCC    &TOME   &GP-MGLA    &WEAC   &WCT    &EAC    &HBGF   &MCLA   &HGPA   &CSPA\\
\midrule
LWEA	&(\textbf{30}$/$0$/$0)	&(\textbf{30}$/$0$/$0)	&(\textbf{24}$/$2$/$4)	&(\textbf{26}$/$4$/$0)	&(\textbf{28}$/$2$/$0)	&(\textbf{28}$/$2$/$0)	&(\textbf{29}$/$1$/$0)	&(\textbf{27}$/$3$/$0)	&(\textbf{28}$/$2$/$0)	&(\textbf{30}$/$0$/$0)	&(\textbf{29}$/$1$/$0)\\
\midrule
LWGP	&(\textbf{30}$/$0$/$0)	&(\textbf{30}$/$0$/$0)	&(\textbf{23}$/$4$/$3)	&(\textbf{29}$/$1$/$0)	&(\textbf{25}$/$4$/$1)	&(\textbf{26}$/$1$/$3)	&(\textbf{26}$/$3$/$1)	&(\textbf{28}$/$2$/$0)	&(\textbf{27}$/$2$/$1)	&(\textbf{30}$/$0$/$0)	&(\textbf{28}$/$2$/$0)\\
\bottomrule
\end{tabular}
\vskip -0.08in
\end{table*}

The parameter $\theta$ controls the influence of the cluster uncertainty over the consensus process of LWEA and LWGP. A smaller $\theta$ leads to a stronger influence of cluster uncertain over the consensus process via the ECI measure (see Fig.~\ref{fig:uncertainty_vs_eci}).

We evaluate the clustering performances of LWEA and LWGP with varying parameters $\theta$. For each value of parameter $\theta$, we run the proposed LWEA and LWGP methods 20 times, respectively, with the ensemble of base clusterings randomly drawn from the base clustering pool at each time, and report their average NMI scores with varying parameters $\theta$ in Table~\ref{table:para_weac} and Table~\ref{table:para_LWGP}. As can be seen in Table~\ref{table:para_weac} and Table~\ref{table:para_LWGP}, the proposed LWEA and LWGP methods yield consistent clustering performances with different values of $\theta$ on the benchmark datasets. Empirically, it is suggested that the parameter $\theta$ be set to moderate values, e.g., in the interval of $[0.2,1]$. In the following of this paper, for both LWEA and LWGP, we will use $\theta=0.4$ in all experiments on the benchmark datasets.

\begin{figure*}[!th]
\begin{center}
{\subfigure[\emph{Caltech20}]
{\includegraphics[width=0.381\columnwidth]{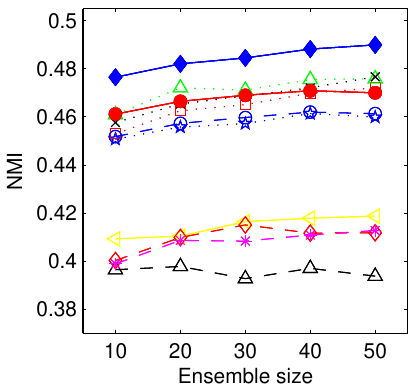}\label{fig:comp_nmi_Msize1}}}
{\subfigure[\emph{FCT}]
{\includegraphics[width=0.381\columnwidth]{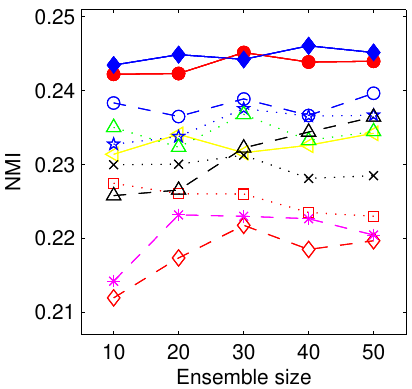}}}
{\subfigure[\emph{IS}]
{\includegraphics[width=0.381\columnwidth]{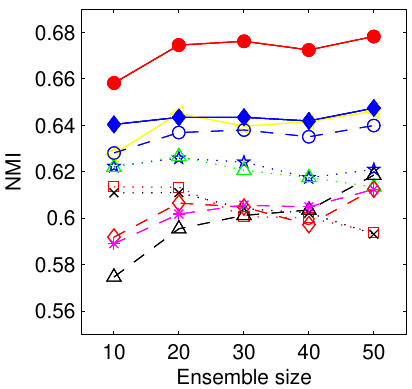}}}
{\subfigure[\emph{ISOLET}]
{\includegraphics[width=0.381\columnwidth]{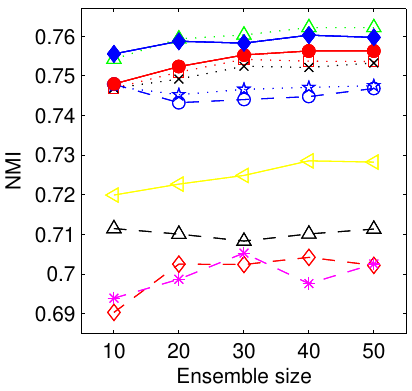}}}
{\subfigure[\emph{LR}]
{\includegraphics[width=0.381\columnwidth]{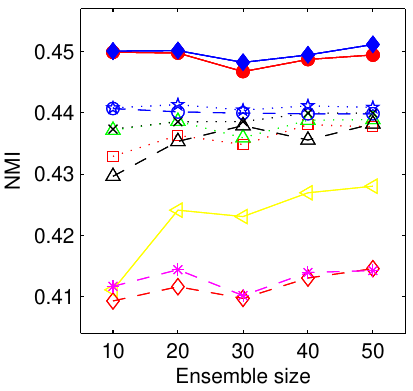}}}
{\subfigure[\emph{LS}]
{\includegraphics[width=0.381\columnwidth]{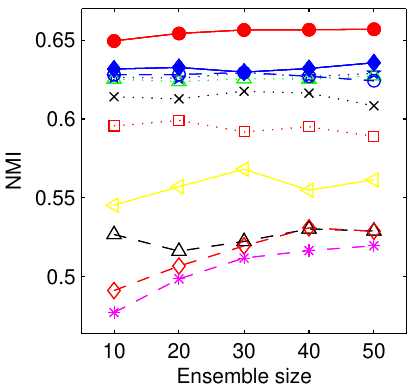}}}
{\subfigure[\emph{MF}]
{\includegraphics[width=0.381\columnwidth]{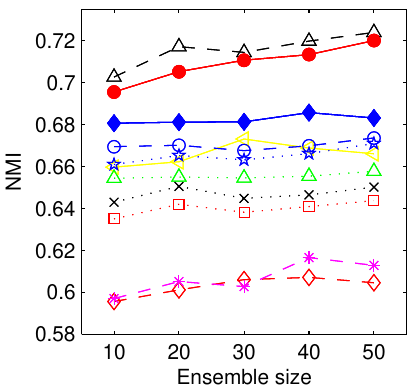}}}
{\subfigure[\emph{MNIST}]
{\includegraphics[width=0.381\columnwidth]{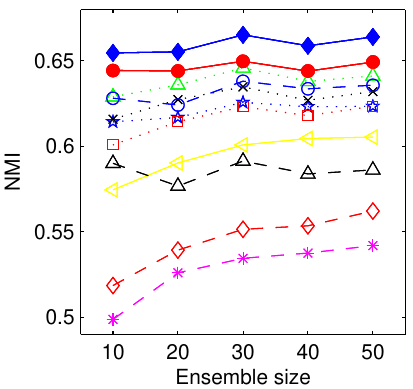}}}
{\subfigure[\emph{ODR}]
{\includegraphics[width=0.381\columnwidth]{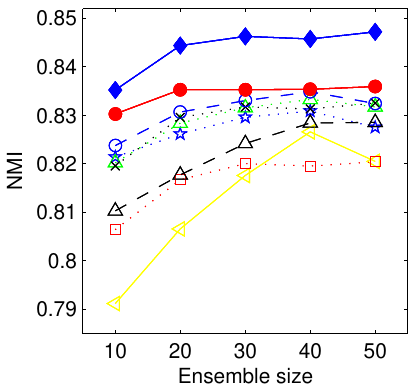}}}
{\subfigure[\emph{PD}]
{\includegraphics[width=0.381\columnwidth]{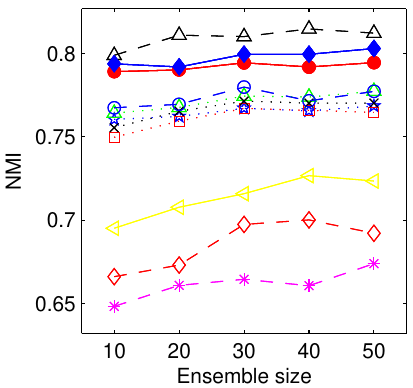}\label{fig:comp_nmi_Msize10}}}
{\subfigure[\emph{Semeion}]
{\includegraphics[width=0.381\columnwidth]{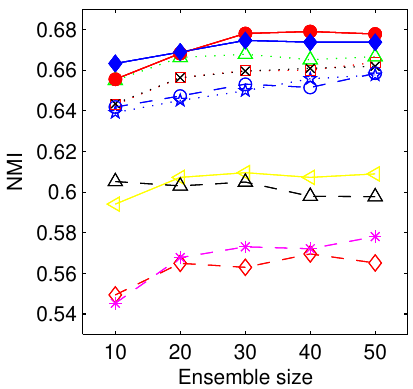}}}
{\subfigure[\emph{SPF}]
{\includegraphics[width=0.381\columnwidth]{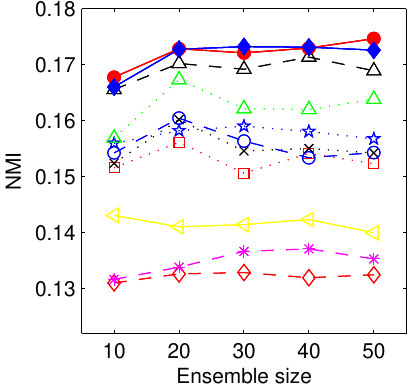}}}
{\subfigure[\emph{Texture}]
{\includegraphics[width=0.381\columnwidth]{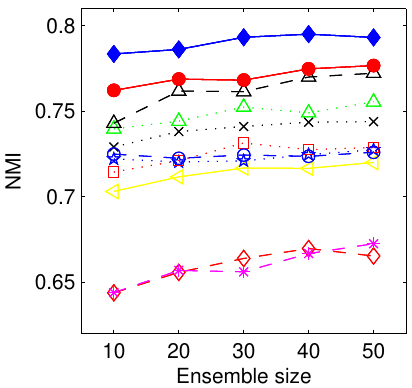}}}
{\subfigure[\emph{VS}]
{\includegraphics[width=0.381\columnwidth]{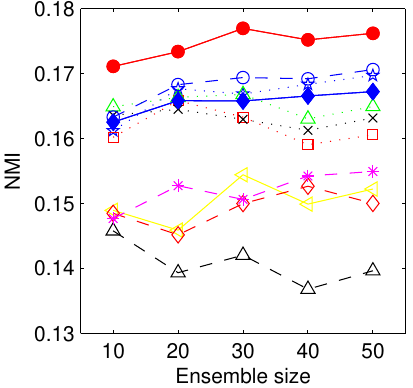}}}
{\subfigure[\emph{USPS}]
{\includegraphics[width=0.381\columnwidth]{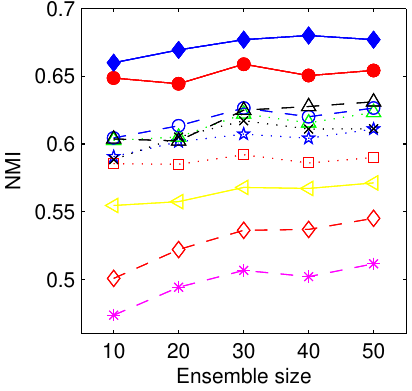}\label{fig:comp_nmi_Msize15}}}
{\subfigure
{\includegraphics[width=1.4\columnwidth]{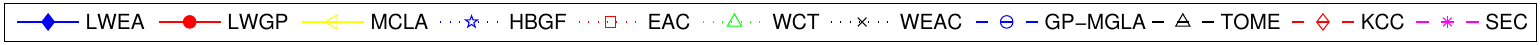}}}\vskip -0.08in
\caption{The average performances (w.r.t. NMI) over 20 runs by different methods with varying ensemble sizes $M$.}\vskip -0.06in
\label{fig:comp_nmi_Msize}\vskip -0.1in
\end{center}
\end{figure*}

\begin{figure*}[!th]
\begin{center}
{\subfigure[\emph{Caltech20}]
{\includegraphics[width=0.381\columnwidth]{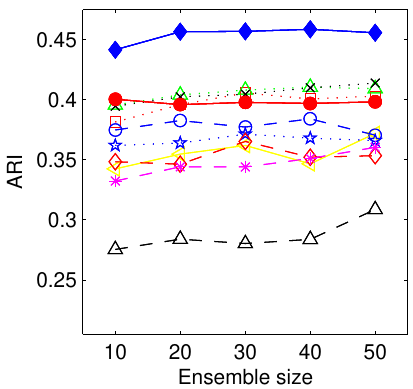}\label{fig:comp_ari_Msize1}}}
{\subfigure[\emph{FCT}]
{\includegraphics[width=0.381\columnwidth]{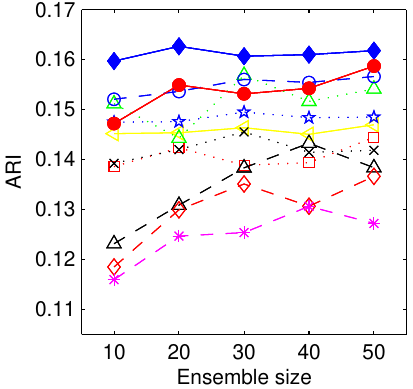}}}
{\subfigure[\emph{IS}]
{\includegraphics[width=0.381\columnwidth]{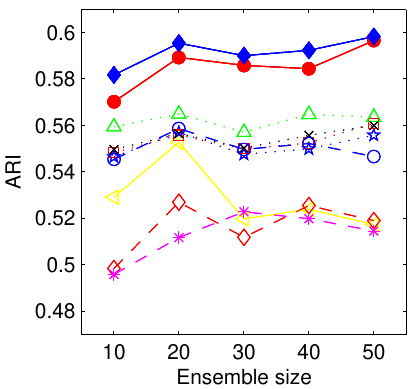}}}
{\subfigure[\emph{ISOLET}]
{\includegraphics[width=0.381\columnwidth]{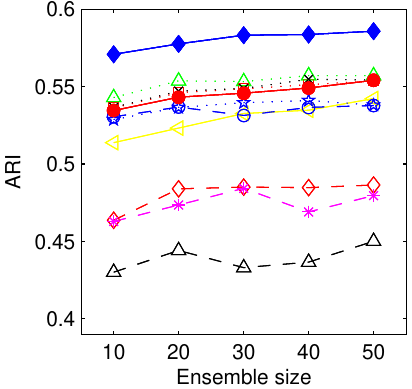}}}
{\subfigure[\emph{LR}]
{\includegraphics[width=0.381\columnwidth]{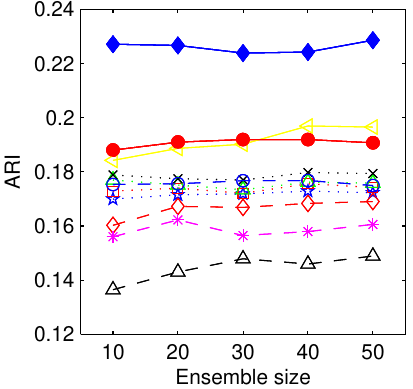}}}
{\subfigure[\emph{LS}]
{\includegraphics[width=0.381\columnwidth]{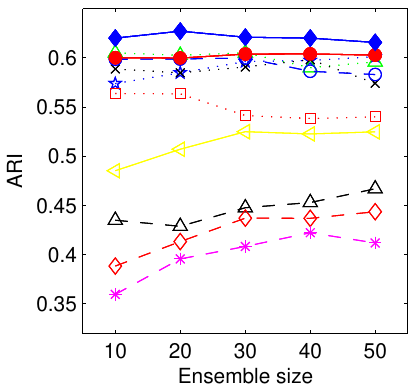}}}
{\subfigure[\emph{MF}]
{\includegraphics[width=0.381\columnwidth]{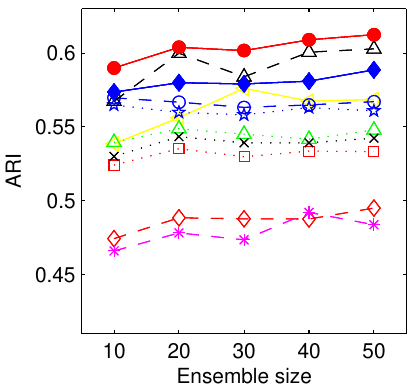}}}
{\subfigure[\emph{MNIST}]
{\includegraphics[width=0.381\columnwidth]{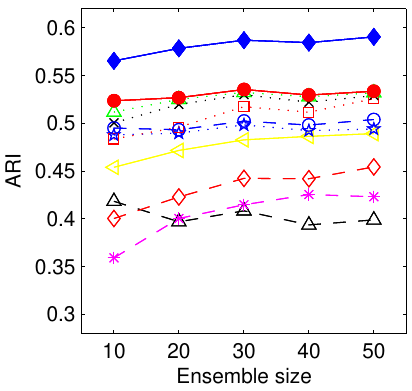}}}
{\subfigure[\emph{ODR}]
{\includegraphics[width=0.381\columnwidth]{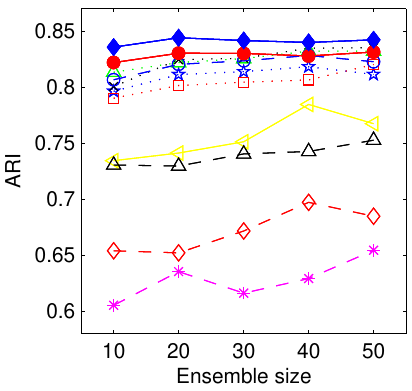}}}
{\subfigure[\emph{PD}]
{\includegraphics[width=0.381\columnwidth]{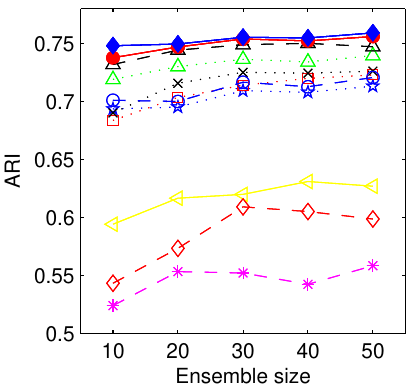}\label{fig:comp_ari_Msize10}}}
{\subfigure[\emph{Semeion}]
{\includegraphics[width=0.381\columnwidth]{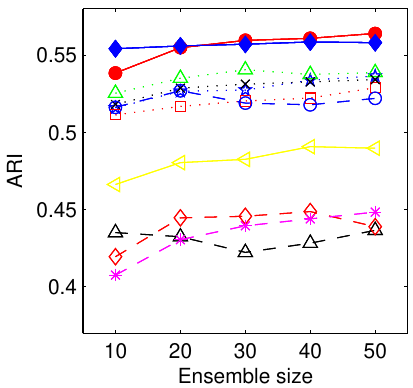}}}
{\subfigure[\emph{SPF}]
{\includegraphics[width=0.381\columnwidth]{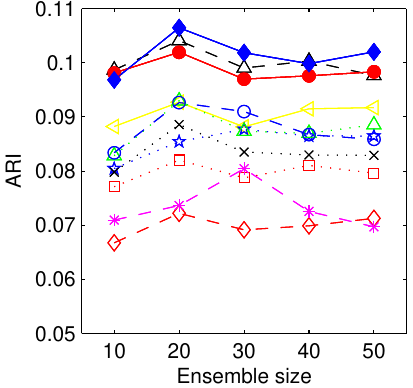}}}
{\subfigure[\emph{Texture}]
{\includegraphics[width=0.381\columnwidth]{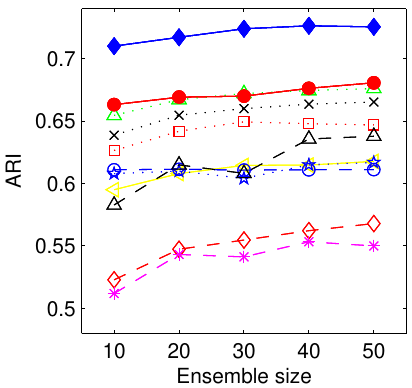}}}
{\subfigure[\emph{VS}]
{\includegraphics[width=0.381\columnwidth]{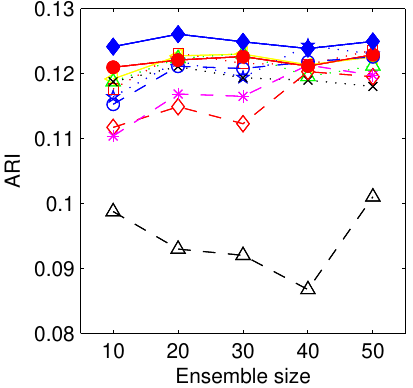}}}
{\subfigure[\emph{USPS}]
{\includegraphics[width=0.381\columnwidth]{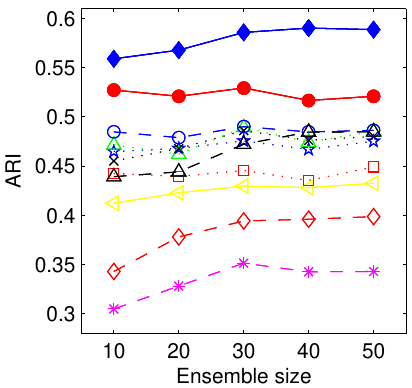}\label{fig:comp_ari_Msize15}}}
{\subfigure
{\includegraphics[width=1.4\columnwidth]{Ensizes_legend}}}\vskip -0.08in
\caption{The average performances (w.r.t. ARI) over 20 runs by different methods with varying ensemble sizes $M$.}\vskip -0.06in
\label{fig:comp_ari_Msize}
\end{center}\vskip -0.1in
\end{figure*}

\subsection{Comparison against Base Clusterings}
\label{comp_base}

The purpose of ensemble clustering is to combine multiple base clusterings to obtain a probably better and more robust consensus clustering. In this section, we compare the consensus clusterings of the proposed LWEA and LWGP methods against the base clusterings. For each benchmark dataset, we run the proposed LWEA and LWGP methods 100 times, respectively, with the ensemble of base clusterings randomly drawn from the pool at each time. The average NMI scores and variances of LWEA, LWGP, as well as the base clusterings are illustrated in Fig.~\ref{fig:base_comp}. The proposed methods exhibit significant improvements over the base clusterings on all the fifteen benchmark datasets (see Fig.~\ref{fig:base_comp}). Especially, for the \emph{IS}, \emph{LS}, \emph{MF}, \emph{MNIST}, \emph{ODR}, \emph{PD}, \emph{Semeion}, \emph{texture}, and \emph{USPS} datasets, the advantage of the proposed methods over the base clusterings is even greater.

\subsection{Comparison against Other Ensemble Clustering Methods}
\label{comp_others}

In this section, we compare the proposed LWEA and LWGP methods against eleven ensemble clustering methods, namely, CSPA \cite{strehl02}, HGPA \cite{strehl02}, MCLA \cite{strehl02}, hybrid bipartite graph formulation (HBGF) \cite{fern04_bipartite}, EAC \cite{Fred05_EAC}, weighted connected triple based method (WCT) \cite{iam_on11_linkbased}, weighted evidence accumulation clustering (WEAC) \cite{huang14_weac}, graph partitioning with multi-granularity link analysis (GP-MGLA) \cite{huang14_weac}, Two-level-refined cO-association Matrix Ensemble (TOME) \cite{Zhong15_pr}, $k$-means based consensus clustering (KCC) \cite{wu15_TKDE}, and spectral ensemble clustering (SEC) \cite{kdd15_sec}. For each of the proposed methods and the baseline methods, we use two criteria to specify the number of clusters for the consensus clustering, that is, best-$k$ and true-$k$. For best-$k$, the number of clusters that leads to the best performance is adopted for each test method. For true-$k$, the actual number of classes in the dataset is adopted for each method.

To achieve a fair comparison, we run each of the proposed methods and the baseline methods 100 times with the ensembles randomly constructed from the base clustering pool (see Section~\ref{sec:dataset_and_eval}). The average performances and standard deviations of different methods over 100 runs are reported in Table~\ref{table:compare_ce_nmi} (w.r.t. NMI) and Table~\ref{table:compare_ce_ari} (w.r.t. ARI).

As shown in Table~\ref{table:compare_ce_nmi}, the proposed LWEA and LWGP methods achieve the best NMI scores on the \emph{IS}, \emph{LR}, \emph{MNIST}, \emph{ODR}, \emph{Semeion}, \emph{Texture}, and \emph{USPS} datasets in terms of both best-$k$ and true-$k$, and nearly the best scores on the \emph{ISOLET}, \emph{LS}, \emph{SPF}, and  \emph{VS} datasets. As shown in Table~\ref{table:compare_ce_ari}, the proposed LWEA and LWGP methods achieve the best ARI scores on the \emph{IS}, \emph{LR}, \emph{ODR}, \emph{Semeion}, \emph{Texture}, and \emph{USPS} datasets in both best-$k$ and true-$k$, and nearly the best ARI scores on the \emph{FCT}, \emph{ISOLET}, \emph{LS}, \emph{MF}, \emph{MNIST}, and \emph{PD} datasets. Although the TOME method outperforms the proposed methods on the \emph{MF} and \emph{PD} datasets w.r.t. NMI, yet on all of the other thirteen datasets it shows a lower or significantly lower NMI scores than our methods (see Table~\ref{table:compare_ce_nmi}). That is probably due to the fact that the TOME method exploits Euclidian distances between objects to improve the consensus process and its efficacy heavily relies on some implicit assumptions on the data distribution, which places an unstable factor for the consensus performance of TOME. To summarize, as shown in Tables~\ref{table:compare_ce_nmi} and \ref{table:compare_ce_ari}, in comparison with the eleven baseline methods, the proposed LWEA and LWGP methods yield overall the best performance on the benchmark datasets.

\begin{figure}[!t]
\begin{center}
{
{\includegraphics[width=0.75\linewidth]{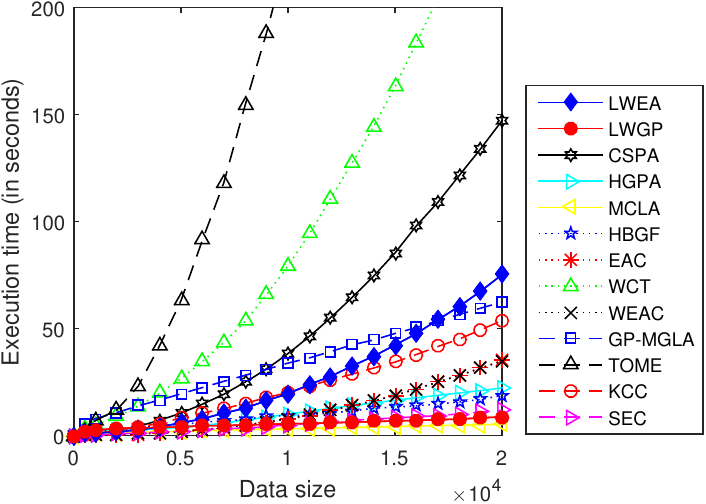}}}\vskip -0.1in
\caption{Execution time of different methods with varying data sizes.}\label{fig:comp_time}
\end{center}\vskip -0.1in
\end{figure}

To further analyze the experimental results in Tables~\ref{table:compare_ce_nmi} and \ref{table:compare_ce_ari}, we use the t-test \cite{dem06_jmlr} (with $p<0.05$) to evaluate the statistical significance of the differences between our methods and the baseline methods. Because fifteen benchmark datasets are used in our experiment and for each dataset we conduct two comparisons (in terms of best-$k$ and true-$k$ respectively), so there are totally 30 comparisons in Table~\ref{table:compare_ce_nmi} and in Table~\ref{table:compare_ce_ari}. It is noteworthy that in each comparison every test method is performed 100 times and their average performances and standard deviations are reported. In a comparison, if our method achieves a higher (or lower) score than a baseline method \emph{and} the difference is statistically significant according to t-test with $p<0.05$, then we say our method is \emph{significantly better} (or \emph{significantly worse}) than the baseline method for one time. If the difference between our method and a baseline method is not statistically significant in a comparison, then we say these two methods are comparable to each other for one time. Table~\ref{table:compare_ce2_nmi} and Table~\ref{table:compare_ce2_ari} report the number of times that the proposed methods are significantly better than \emph{or} comparable to \emph{or} significantly worse than a baseline method w.r.t. NMI and ARI, respectively. Specifically, as shown in Table~\ref{table:compare_ce2_nmi}, in terms of NMI, the proposed LWEA and LWGP methods exhibit statistically significant improvements over the SEC, KCC, and HGPA methods in all of the 30 comparisons, and statistically significantly outperform each of the other eight baseline methods at least 23 times out of the totally 30 comparisons. Similar advantages can also be observed in Table~\ref{table:compare_ce2_ari}, which shows that LWEA and LWGP significantly outperform each baseline method (w.r.t. ARI) at least 23 times out of the totally 30 comparisons according to t-test.

\subsection{Robustness to Ensemble Sizes $M$}
\label{sec:ensize}

Furthermore, we evaluate the performances of our methods and the baseline methods with varying ensemble sizes $M$. For each ensemble size $M$, we run the proposed methods and the baseline methods 20 times on each benchmark dataset, with the ensemble of $M$ base clusterings randomly selected at each time. Then we illustrate the average performances, w.r.t. NMI and ARI, of different methods with varying ensemble sizes in Fig.~\ref{fig:comp_nmi_Msize} and Fig.~\ref{fig:comp_ari_Msize}, respectively. In terms of NMI, the TOME method yields better performance than the proposed methods in the \emph{MF} and \emph{PD} datasets, but in all of the other thirteen datasets the proposed methods significantly outperform the TOME method. As shown in Fig.~\ref{fig:comp_nmi_Msize}, compared with the baseline methods, the proposed methods achieve overall the most consistent and robust performances (w.r.t. NMI) with varying ensemble sizes on the benchmark datasets. When it comes to the ARI measure, the proposed LWEA and LWGP methods still achieve the best or nearly the best ARI scores on each benchmark dataset and exhibit overall the best performances with varying ensemble sizes (as shown in Fig.~\ref{fig:comp_ari_Msize}).

\subsection{Execution Time}
\label{sec:time}

In this section, we compare the execution time of different ensemble clustering methods with varying data sizes. The experiments are performed on different subsets of the \emph{LR} dataset. The \emph{LR} dataset consists of totally $20,000$ data objects. When testing the data size of $N'$, we randomly select a subset of $N'$ objects from the \emph{LR} dataset and run different methods on this subset to evaluate their execution time. As illustrated in Fig.~\ref{fig:comp_time}, the proposed LWEA method requires 75.20 seconds to process the entire \emph{LR} dataset, which is comparable to GP-MGLA but much faster than CSPA, WCT, SRS, and TOME. Out of the totally thirteen test methods, the MCLA method is the fastest method, while the proposed LWGP method is the second fastest method. The MCLA method and the proposed LWGP method consume 5.31 seconds and 8.74 seconds respectively to process the entire \emph{LR} dataset. Note that, although the proposed LWGP method is slightly slower than MCLA (but faster than all of the other eleven test methods), yet it significantly outperforms MCLA in clustering accuracy and robustness on the benchmark datasets (see Tables~\ref{table:compare_ce_nmi}, \ref{table:compare_ce2_nmi}, \ref{table:compare_ce_ari}, and \ref{table:compare_ce2_ari} and Figs.~\ref{fig:comp_nmi_Msize} and \ref{fig:comp_ari_Msize}).

To summarize, as shown in the experimental results on various datasets (see Tables~\ref{table:compare_ce_nmi}, \ref{table:compare_ce2_nmi}, \ref{table:compare_ce_ari}, and \ref{table:compare_ce2_ari} and Figs.~\ref{fig:comp_nmi_Msize}, \ref{fig:comp_ari_Msize}, and \ref{fig:comp_time}), the proposed LWEA and LWGP methods yield more consistent and better consensus performances than the baseline methods while exhibiting competitive efficiency.

All experiments are conducted in MATLAB R2014a 64-bit on a workstation (Windows Server 2008 R2 64-bit, 8 Intel 2.40 GHz processors, 96 GB of RAM).

\section{Conclusion}
\label{sec:conclusion}
In this paper, we have proposed a novel ensemble clustering approach based on ensemble-driven cluster uncertainty estimation and local weighting strategy. We propose to estimate the uncertainty of clusters by considering the cluster labels in the entire ensemble based on an entropic criterion, and devise a new ensemble-driven cluster validity index termed ECI. The ECI measure requires no access to the original data features and makes no assumptions on the data distribution. Then, a local weighting scheme is presented to extend the conventional CA matrix into the LWCA matrix via the ECI measure. With the reliability of clusters investigated and the local diversity in ensembles exploited, we further propose two novel consensus functions, termed LWEA and LWGP, respectively. We have conducted extensive experiments on a variety of real-world datasets. The experimental results have shown the superiority of the proposed approach in terms of both clustering quality and efficiency when compared to the state-of-the-art approaches.

\ifCLASSOPTIONcaptionsoff
  \newpage
\fi

\bibliographystyle{IEEEtran}

\bibliography{tcyb_2016}

\begin{IEEEbiography}[{\includegraphics[width=1in,height=1.25in,clip,keepaspectratio]{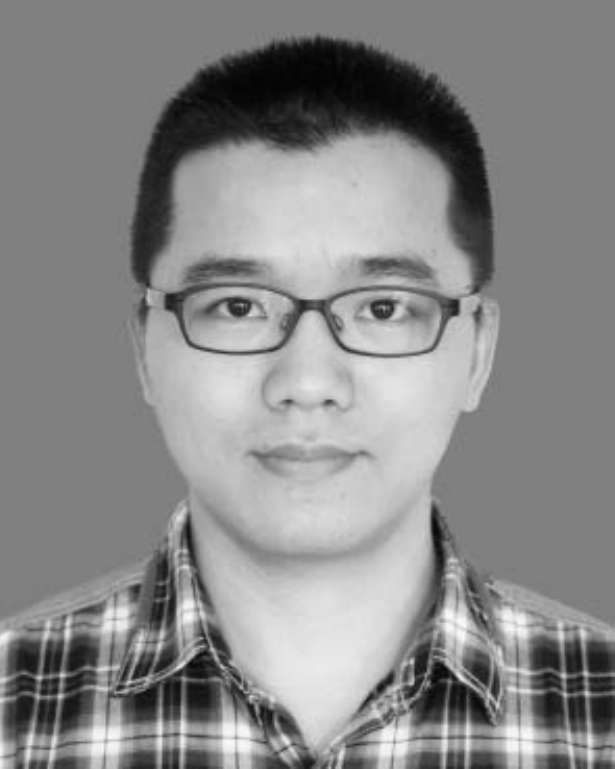}}]{Dong Huang}
received his B.S. degree in computer science in 2009 from South China University of Technology, China. He received his M.Sc. degree in computer science in 2011 and his Ph.D. degree in computer science in 2015, both from Sun Yat-sen University, China. He joined South China Agricultural University in 2015, where he is currently an Associate Professor with the College of Mathematics and Informatics. His research interests include data mining and pattern recognition. He is a member of the IEEE.
\end{IEEEbiography}

\begin{IEEEbiography}[{\includegraphics[width=1in,height=1.25in,clip,keepaspectratio]{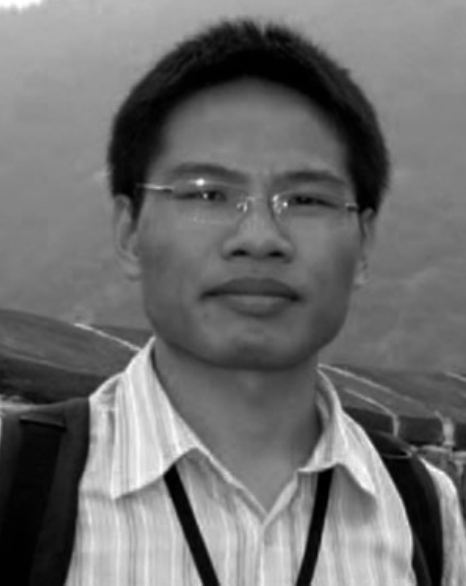}}]{Chang-Dong Wang}
received the B.S. degree in applied mathematics in 2008, the M.Sc. degree in computer science in 2010, and the Ph.D. degree in computer science in 2013, all from Sun Yat-sen University, Guangzhou, China. He was a visiting student at the University of Illinois at Chicago from January 2012 to November 2012. He is currently an Assistant Professor with the School of Data and Computer Science, Sun Yat-sen University, Guangzhou, China. His current research interests include machine learning and data mining. He has published more than 40 scientific papers in international journals and conferences such as IEEE TPAMI, IEEE TKDE, IEEE TSMC-C, Pattern Recognition, KAIS, Neurocomputing, ICDM and SDM. His ICDM 2010 paper won the Honorable Mention for Best Research Paper Awards. He was awarded 2015 Chinese Association for Artificial Intelligence (CAAI) Outstanding Dissertation. He is a member of the IEEE.
\end{IEEEbiography}

\begin{IEEEbiography}[{\includegraphics[width=1in,height=1.25in,clip,keepaspectratio]{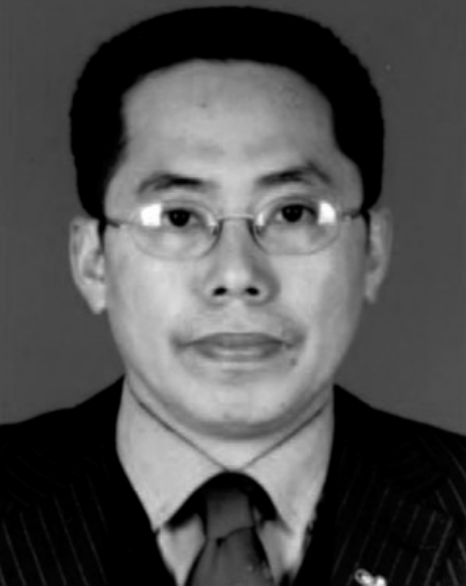}}]{Jian-Huang Lai}
received the M.Sc. degree in applied mathematics in 1989 and the Ph.D. degree in mathematics in 1999 from Sun Yat-sen University, China. He joined Sun Yat-sen University in 1989 as an Assistant Professor, where he is currently a Professor with the School of Data and Computer Science. His current research interests include the areas of digital image processing, pattern recognition, multimedia communication, wavelet and its applications. He has published more than 200 scientific papers in the international journals and conferences on image processing and pattern recognition, such as IEEE TPAMI, IEEE TKDE, IEEE TNN, IEEE TIP, IEEE TSMC-B, Pattern Recognition, ICCV, CVPR, IJCAI, ICDM and SDM. Prof. Lai serves as a Standing Member of the Image and Graphics Association of China, and also serves as a Standing Director of the Image and Graphics Association of Guangdong. He is a senior member of the IEEE.
\end{IEEEbiography}

\end{document}